\documentclass[10pt,twocolumn,letterpaper]{article}

\usepackage[pagenumbers]{cvpr} % To force page numbers, e.g. for an arXiv version

\usepackage{multirow}
\usepackage{color}
\usepackage{balance}
\usepackage{cases}
\usepackage{array}
\usepackage{bbding}
\usepackage{amssymb}
\usepackage{graphicx}
\usepackage{booktabs}
\usepackage{diagbox}
\usepackage{caption}

\usepackage{colortbl} 
\usepackage{amsmath}
\usepackage{empheq}

\usepackage{algorithm}  
\usepackage{algorithmic}  

\usepackage[dvipsnames]{xcolor}

\definecolor{cvprblue}{rgb}{0.51,0.79,0.54}
\definecolor{cvprgray}{RGB}{250,235,235}
\definecolor{cvprgray2}{RGB}{255, 250, 240}
\definecolor{cvprblue2}{RGB}{50, 100, 220}
\definecolor{cvprred}{RGB}{240,100,80}
\usepackage[pagebackref,breaklinks,colorlinks,citecolor=cvprblue]{hyperref}

\def\paperID{3634} % *** Enter the Paper ID here
\def\confName{CVPR}
\def\confYear{2026}

\title{Beyond Single Solution: Multi-Hypothesis Collaborative Deep Unfolding \\ Network for Image Compressive Sensing}

\author{
Wenxue Cui$^{1}$ \ \
Hualin Li$^{1}$ \ \
Yuhang Qin$^{1}$ \ \
Yifu Xu$^{1}$ \ \
Xiaopeng Fan$^{1,2}$ \thanks{Corresponding author.} \ \
Debin Zhao$^{1}$  \\
\normalsize $^{1}$Harbin Institute of Technology, Harbin, China \\
\normalsize $^{2}$Harbin Institute of Technology Suzhou Research Institute, Suzhou, China \\
{\tt\small wxcui@hit.edu.cn \quad 25S103461@stu.hit.edu.cn \quad 120L022221@stu.hit.edu.cn} \\
\vspace{-0.1in}
{\tt\small xuyifu@stu.hit.edu.cn \quad fxp@hit.edu.cn \quad dbzhao@hit.edu.cn} 
}

\begin{document}
\maketitle

\begin{abstract}

Recent deep unfolding networks (DUNs) have advanced Compressive Sensing (CS) by effectively integrating iterative optimization with deep learning architectures. However, most CS approaches predominantly confine their inference to a single solution space, neglecting the inherent ill-posedness of CS problems that intrinsically permits multiple plausible candidate hypotheses. In this paper, a novel Multi-Hypothesis Collaborative Deep Unfolding CS Network (MHC-DUN) is proposed, which explicitly models and leverages multiple hypotheses by jointly optimizing across diverse solution spaces. Specifically, following the Proximal Gradient Descent algorithm, MHC-DUN jointly performs gradient descent and proximal mapping within this multi-hypothesis paradigm. \textbf{i)} For gradient descent, a well-designed AlphaNet is introduced to dynamically predict spatially varying step sizes for all hypotheses, enabling collaborative gradient updates across multiple solutions. \textbf{ii)} For proximal operator, a sophisticated multi-hypothesis collaborative proximal mapping module is designed, which leverages both intra-hypothesis and inter-hypothesis correlation priors to jointly refine multiple solutions. To enable end-to-end training, a novel composite loss function is designed, which balances measurement fidelity, hypothesis diversity, and reconstruction accuracy, encouraging exploration of complementary solutions while maintaining reconstruction fidelity. Experimental results reveal that the proposed CS method outperforms existing CS networks.
\end{abstract}

\vspace{-0.1in}
\section{Introduction}
\label{sec:intro}

% Compressive Sensing (CS)~\cite{donoho2006compressed, candes2008introduction}, an emerging paradigm for signal acquisition, has garnered considerable attention recently in the field of image processing. Specifically, the CS theory posits that if a signal exhibits sparsity in a certain domain, it can be accurately reconstructed from a substantially smaller number of linear measurements than mandated by the Nyquist sampling theorem \cite{liutkus2014imaging}. By enabling the simultaneous sampling and compression, CS provides a groundbreaking approach to efficient data acquisition and reconstruction. As above, CS has been widely applied across various fields, such as single-pixel cameras~\cite{4472247,7778203}, magnetic resonance imaging (MRI)~\cite{lustig2007sparse,6488855}, and snapshot compressive imaging~\cite{8481592,9363502}.

Compressive Sensing (CS)~\cite{donohoCompressedSensing2006, candesIntroductionCompressiveSampling2008}, an emerging paradigm for signal acquisition, has attracted significant interest within the image processing community. Specifically, CS theory posits that a signal exhibiting sparsity in a certain domain can be accurately reconstructed from far fewer linear measurements than those mandated by the Nyquist sampling theorem~\cite{liutkusImagingNatureCompressive2014}. By enabling simultaneous sampling and compression, CS offers a groundbreaking approach to data acquisition, enabling a variety of applications including single-pixel cameras~\cite{duarteSinglepixelImagingCompressive2008,roussetAdaptiveBasisScan2017}, magnetic resonance imaging (MRI)~\cite{lustigSparseMRIApplication2007,lingalaBlindCompressiveSensing2013}, snapshot compressive imaging~\cite{liuRankMinimizationSnapshot2019,yuanSnapshotCompressiveImaging2021}.

% Compressive sensing (CS), as a powerful technique for signal acquisition that performs signal sampling and compression simultaneously, has attracted much attention over the past few years. The CS theory depicts that if a signal is sparse in a certain domain, it can be reconstructed from much fewer linear measurements than that suggested by the Nyquist sampling theorem. Due to the simple and fast sampling procedure, the CS technique has achieved great success in many image systems, including magnetic resonance imaging (MRI), single-pixel cameras, and snapshot compressive imaging.

\begin{figure}[t]
\begin{center}

% \hspace{1.06in}
\includegraphics[width=1.05in]{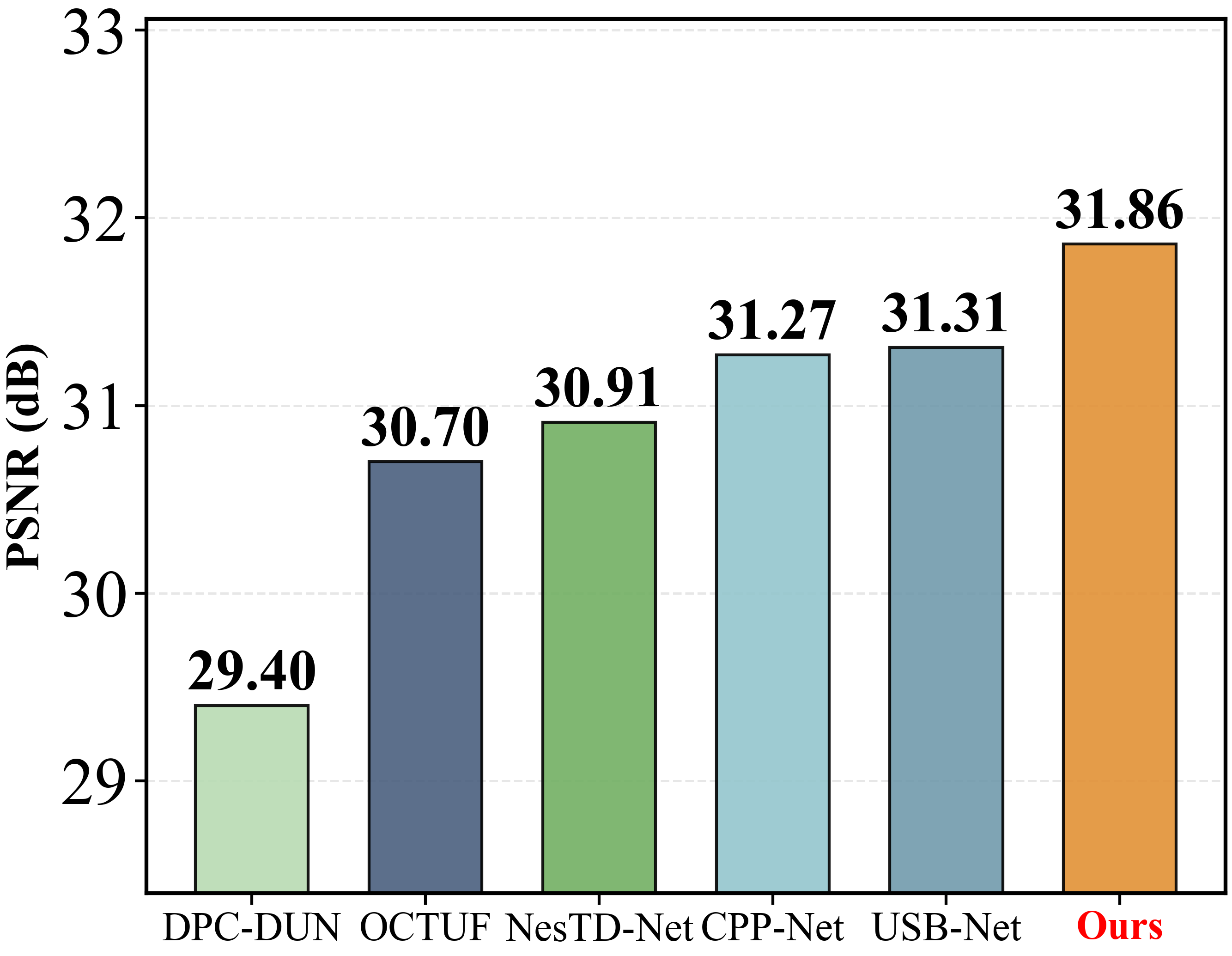}  \hspace{-0.001in}
\includegraphics[width=1.05in]{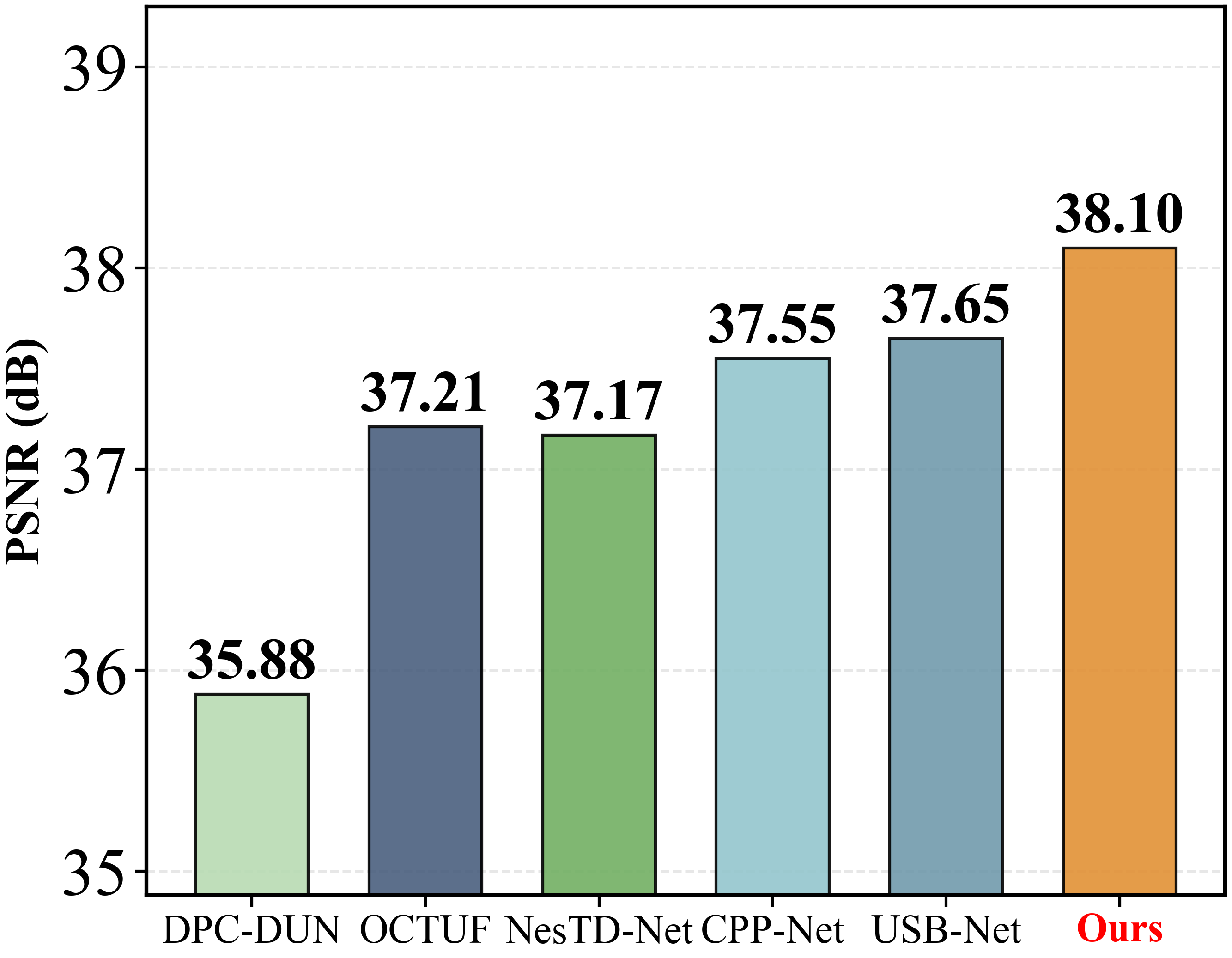}  \hspace{-0.001in}
\includegraphics[width=1.05in]{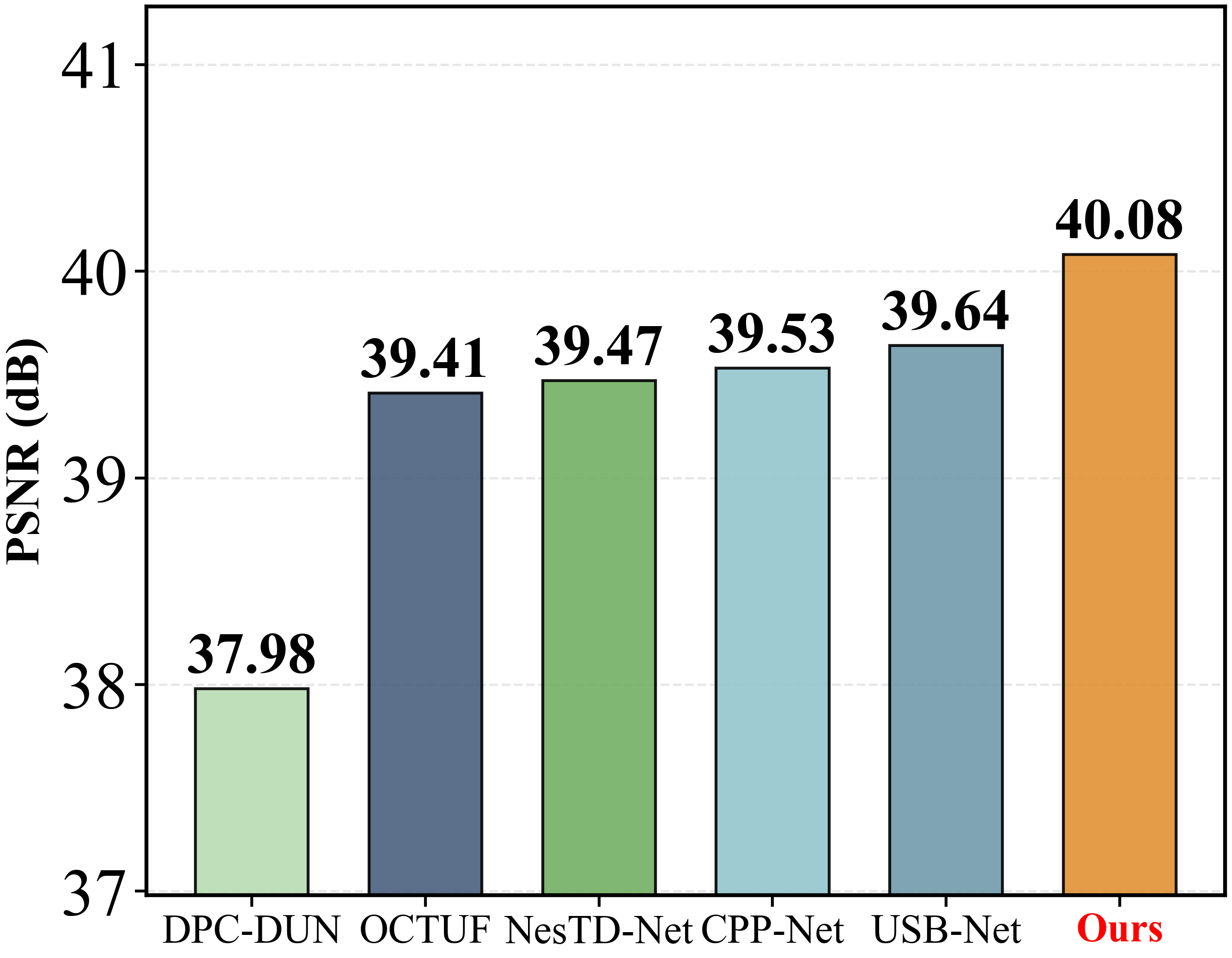}  \hspace{-0.001in}
\\
\vskip -0.18in \tiny{\quad\ \quad R=0.10 on Set11\quad\quad\quad\quad\quad\quad\ \quad\ \quad\quad R=0.30 on Set11\quad\ \quad\quad\ \quad\quad\quad\quad\quad\quad R=0.40 on Set11 }
\vskip 0.02in
\includegraphics[width=1.05in]{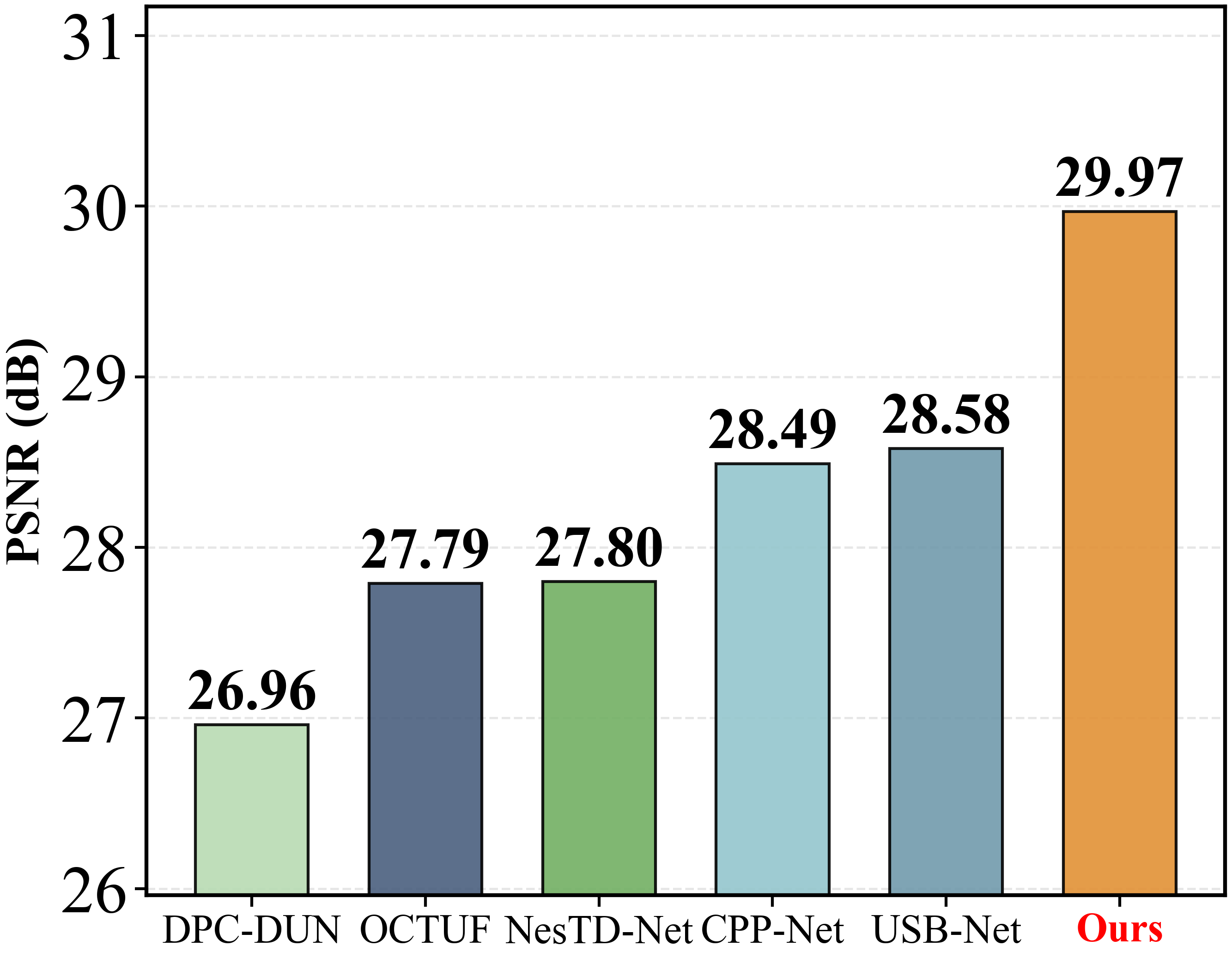}  \hspace{0.037in}
\includegraphics[width=1.05in]{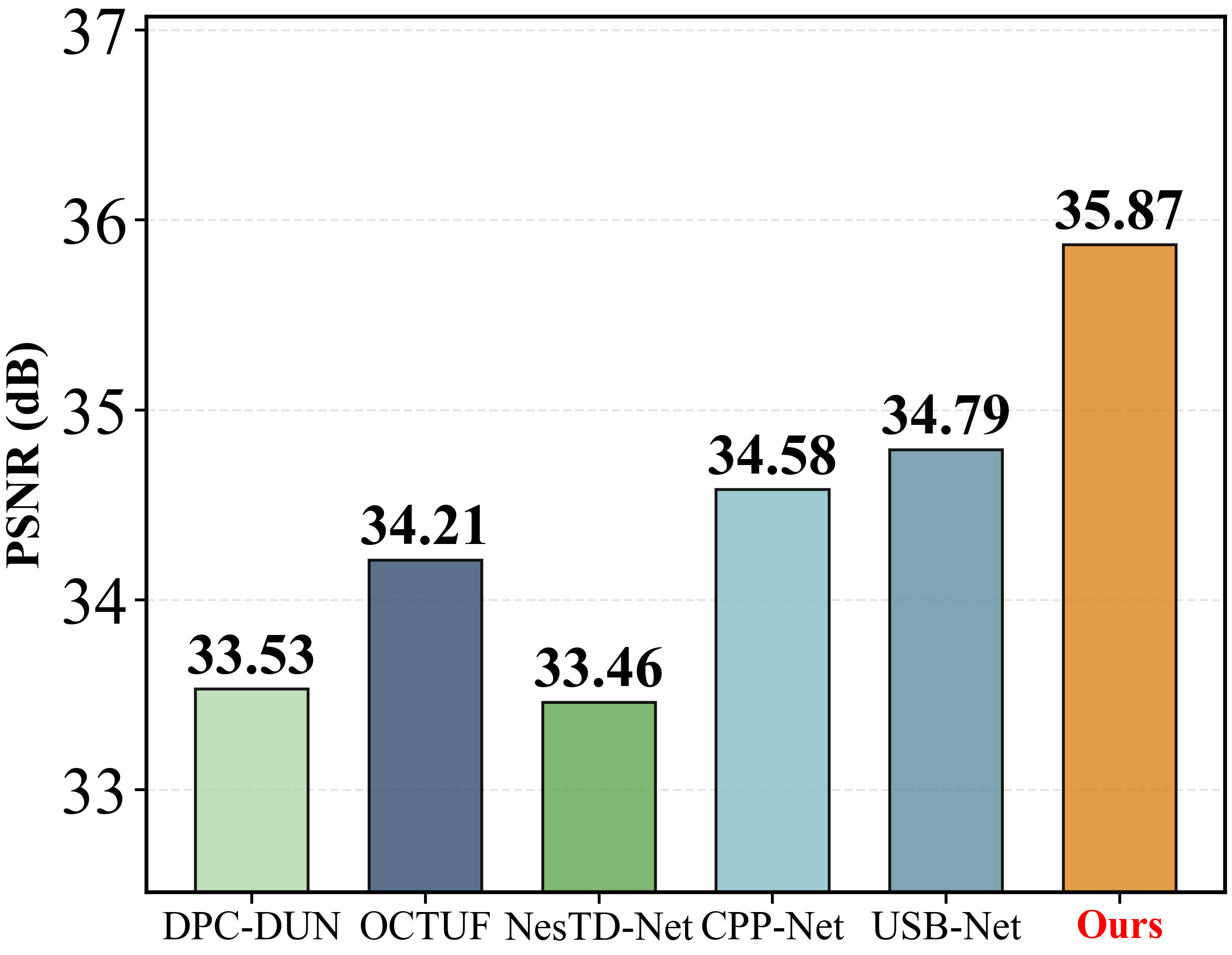}  \hspace{0.029in}
\includegraphics[width=1.05in]{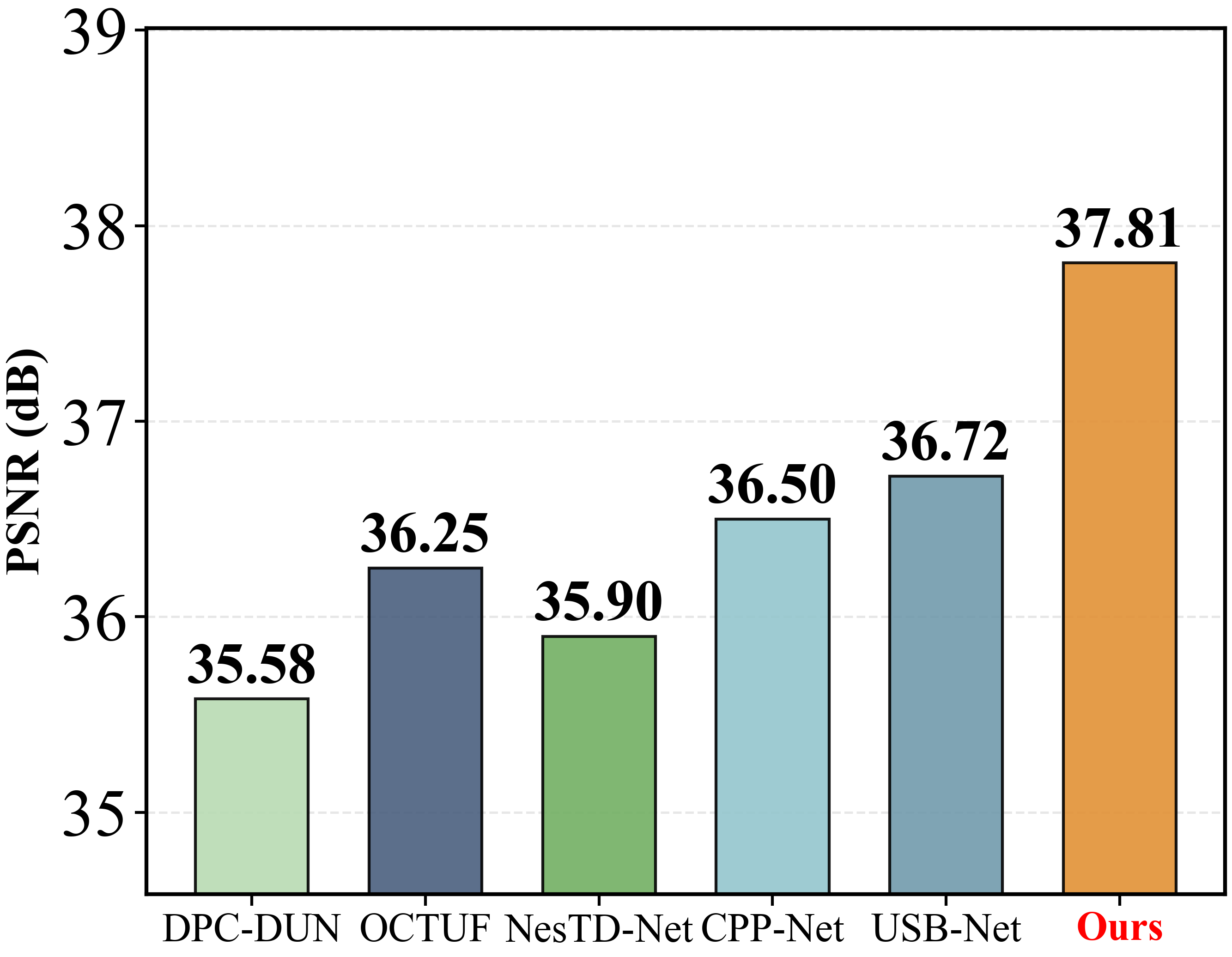}  % \hspace{-0.001in}
\vskip 0.0in \tiny{\quad\quad\  R=0.10 on Urban100\quad\quad\quad\quad\ \quad\ \quad\quad R=0.30 on Urban100\quad\quad\ \quad\ \quad\quad\quad\quad R=0.40 on Urban100 }

\end{center}
\vskip -0.22in
\caption{Performance comparison on PSNR(dB) of the proposed MHC-DUN with the state-of-the-art methods DPC-DUN~\cite{songDynamicPathControllableDeep2023}, OCTUF~\cite{songOptimizationInspiredCrossAttentionTransformer2023}, NesTD-Net~\cite{ganNesTDnetDeepNESTAinspired2024}, CPP-Net~\cite{guoCPPNetEmbracingMultiScale2024}, USB-Net~\cite{guoUSBNetUnfoldingSplit2025}. Our approach can surpass the existing methods by 0.4dB$\sim$1.4dB.}
\vskip -0.21in
\label{fig:1}
\end{figure}

Given the input signal $\mathbf{x}\in\mathbb{R}^{N}$, the sampling process can be represented as $\mathbf{y}=\mathbf{A}\mathbf{x}$, where $\mathbf{y}\in\mathbb{R}^{M}$ denotes the measurement vector acquired through the sampling matrix  $\mathbf{A}\in\mathbb{R}^{M\times N}$, and $M/N$ represents the CS sampling ratio. Since $M\ll N$, recovering $\mathbf{x}$ from $\mathbf{y}$ constitutes an ill-posed inverse problem~\cite{zhangOptimizationInspiredCompactDeep2020}, which typically requires the incorporation of prior knowledge about the signal to effectively regularize the solution space. As above, the corresponding optimization model can be formulated as:
\vskip -0.1in
\begin{equation}
\tilde{\mathbf{x}} = \arg\min_{\mathbf{x}} \frac{1}{2} \|\mathbf{A} \mathbf{x} - \mathbf{y} \|_2^2 + \lambda \mathcal{R}(\mathbf{x}).
\label{e1}
\end{equation}
where the data-fidelity term $\frac{1}{2} \|\mathbf{A} \mathbf{x} - \mathbf{y} \|_2^2$ ensures consistency with the observed measurements, and the regularization term $\mathcal{R}(\mathbf{x})$ encodes prior knowledge about the signal. The parameter $\lambda > 0$ controls the trade-off between these two components. To solve the problem in Eq.~\eqref{e1}, a variety of structure-inducing regularizers~\cite{kimCompressedSensingUsing2007,chenCompressedsensingRecoveryImages2011} have been employed to leverage inherent data priors. These include sparsity in transform domains (e.g., DCT~\cite{zhaoImageCompressivesensingRecovery2014} and wavelet~\cite{anselmiWaveletBasedCompressiveImaging2015}), low-rank structure~\cite{dongCompressiveSensingNonlocal2014,golbabaeeHyperspectralImageCompressed2012}, and non-local self-similarity~\cite{zhaGroupSparsityResidual2020,zhangGroupBasedSparseRepresentation2014}, among others~\cite{candesRobustUncertaintyPrinciples2006,metzlerLearnedDAMPPrincipled2017}. By incorporating these image priors, such optimization-based methods have achieved substantial performance. Nevertheless, these CS algorithms often suffer from considerable computational overhead due to their iterative nature.

Fueled by the potent learning capabilities of deep neural networks, numerous deep learning-based CS approaches have emerged. From an interpretability standpoint, existing CS networks can be broadly categorized into two classes: Deep Black-box Networks (DBNs) and Deep Unfolding Networks (DUNs).
\textbf{1)} DBNs~\cite{yaoDR2NetDeepResidual2019,kulkarniReconNetNonIterativeReconstruction2016,shenMTCCSNetMarryingTransformer2024,sunDualPathAttentionNetwork2020} usually establish a direct, black-box mapping from compressed measurements to the original signal. As a pioneering paradigm, their simplicity and efficiency drove extensive investigation in early deep learning-based CS research. However, this paradigm often lacks theoretical rigor, compromising interpretability and potentially limiting reconstruction fidelity.
\textbf{2)} In contrast, DUNs~\cite{cuiImageCompressedSensing2023,zhangAMSNetAdaptiveMultiScale2022,wangUFCNetUnrollingFixedpoint2024,songOptimizationInspiredCrossAttentionTransformer2023} derive their interpretability by unfolding established optimization solvers (e.g., ISTA~\cite{beckFastIterativeShrinkageThresholding2009}, HQS~\cite{zhangLearningDeepCNN2017}, and AMP~\cite{wangVersatileDenoisingbasedApproximate2023}) into deep architectures. Specifically, inspired by iterative optimization, DUNs usually employ cascaded multi-stage structures for progressive signal reconstruction. By design, this unfolding approach provides DUNs with robust theoretical foundations and enhanced interpretability. However, most existing DUNs still face the following challenges:
\begin{enumerate}[label=\textbf{\alph*.}]
	\item{Due to the inherent ill-posedness of CS systems, which admit multiple plausible candidate solutions, the conventional strategy of directly performing inference in a single solution space usually exacerbates the challenge of regressing optimal results, thereby consequently degrading CS reconstruction quality.}
	\item{The existing single-solution-space inference paradigm generally conducts information extraction and integration within a single feature domain, thereby overlooking the intrinsic correlations among multiple potential solutions, thereby limiting effective interaction across the corresponding feature domains.}
\end{enumerate}

To address above limitations, we propose MHC-DUN, a novel multi-hypothesis collaborative deep unfolding network for image CS reconstruction that explicitly models and exploits multiple candidate solutions via joint optimization over diverse solution manifolds. Specifically, in the gradient descent step, we introduce AlphaNet, a dedicated module that adaptively predicts spatially varying step sizes across all hypotheses, enabling synchronized gradient updates among multiple solutions. In the proximal mapping step, we design an advanced Multi-Hypothesis Collaborative Block (MHCB) that leverages both intra-hypothesis local priors and inter-hypothesis correlation dependencies to jointly refine candidate reconstructions. Finally, to facilitate end-to-end training, we formulate a composite loss function that balances data fidelity, solution diversity, and perceptual quality, encouraging the exploration of plausible reconstructions while maintaining high accuracy. As shown in Fig.~\ref{fig:1}, the experimental results demonstrate that our proposed MHC-DUN significantly outperforms the existing state-of-the-art CS methods.

The main contributions are summarized as follows:

\begin{itemize}[leftmargin=2em]

\item A novel Multi-Hypothesis Collaborative Deep Unfolding CS Network (dubbed MHC-DUN) is proposed, which explicitly represents and exploits multiple plausible hypotheses by performing joint optimization across a range of solution spaces.

\item For the gradient descent, a specialized AlphaNet is introduced to adaptively generate fine-grained, spatially varying step-size maps for each hypothesis, thereby facilitating coordinated dynamic gradient updates across multiple solution candidates.

\item For the proximal operation, a novel multi-hypothesis collaborative proximal mapping network is proposed to jointly refine multiple hypotheses by leveraging both intra-hypothesis structural knowledge and inter-hypothesis correlation priors.

% For the proximal operation, a sophisticated multi-hypothesis collaborative proximal mapping network is developed to jointly refine multiple hypotheses by integrating both intra-hypothesis structural knowledge and inter-hypothesis correlation information.

\item A specially-designed composite loss function is proposed to guide end-to-end training by measuring measurement consistency, hypothesis diversity, and reconstruction quality, encouraging exploration of plausible hypotheses while maintaining reconstruction fidelity.

\end{itemize}

\begin{figure*}[t]
  \centering
  % \fbox{\rule{0pt}{2in} \rule{0.95\linewidth}{0pt}}
  \includegraphics[width=0.95\linewidth]{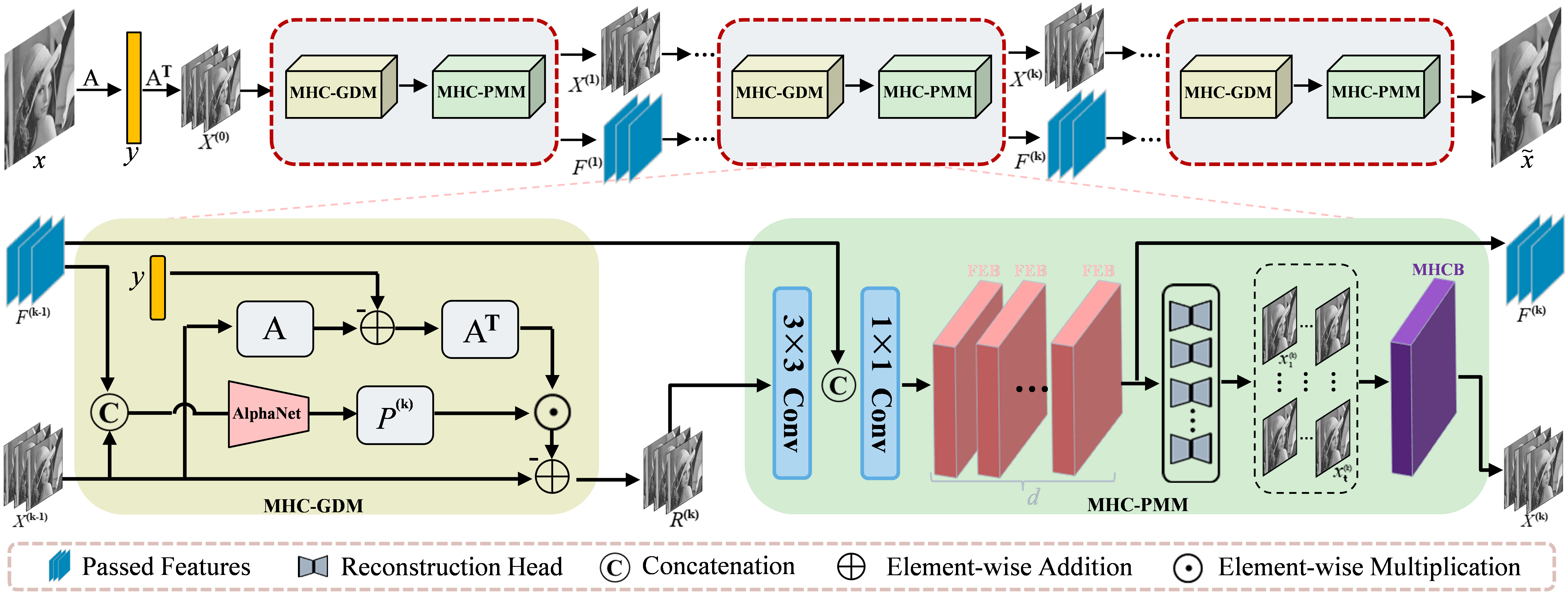}

\vskip -0.1in
   \caption{Architecture of our proposed multi-hypothesis collaboratively deep unfolding network MHC-DUN, which consists of $K$ cascaded stages. Specifically, in each phase, two modules, \ie, Multi-Hypothesis Collaborative Gradient Descent Module (MHC-GDM) and Multi-Hypothesis Collaborative Proximal Mapping Module (MHC-PMM), are included.}
   \label{fig:2}
\vskip -0.18in
\end{figure*}

\section{Related Work}
\label{sec:rw}

\subsection{Deep Black-box Image CS Methods}

Deep black-box image CS methods typically utilize end-to-end networks to reconstruct images directly from measurements. 
Early techniques~\cite{yaoDR2NetDeepResidual2019,metzlerLearnedDAMPPrincipled2017,xieAdaptiveMeasurementNetwork2017} usually perform block-wise sampling and reconstruction, concatenating the outputs to form the final image. 
For example, ReconNet~\cite{kulkarniReconNetNonIterativeReconstruction2016} employs a fully connected layer followed by six convolutional layers for this purpose. 
Lohit \textit{et al.}~\cite{lohit2018convolutional} improved reconstruction quality by substituting the fully connected layer with circulant layers. 
However, these block-wise CS approaches usually suffer from serious blocking artifacts.

To mitigate such artifacts, recent methods~\cite{shiScalableConvolutionalNeural2019,zhouMultiChannelDeepNetworks2021} have shifted towards performing model inference in the global image space. 
For instance, CS-Net~\cite{shiDeepNetworksCompressed2017} concatenates all patches for an initial reconstruction, which is then refined by a CNN-based module that leverages global priors. 
Subsequently, Cui \textit{et al.}~\cite{cuiImageCompressedSensing2023} introduced NL-CSNet, a non-local neural network that employs attention mechanisms to model long-range dependencies. 
More recent Transformer-based approaches, such as TCS-Net~\cite{ganPatchPixelTransformerBased2023} and CSformer~\cite{yeCSformerBridgingConvolution2023}, utilize self-attention modules for progressive, global-context modeling and hierarchical reconstruction. 
Compared to their block-wise counterparts, these global inference strategies effectively suppress blocking artifacts and improve the overall reconstruction quality.

% To address the above problem, some methods \cite{2019Image}, \cite{2019Scalable}, \cite{9159912},  attempt to perform model inference directly in the global image space. For example, CS-Net \cite{shi2017deep} concatenates all image patches in the initial reconstruction process, followed by a deep reconstruction using Convolutional Neural Networks (CNNs), which is able to capture and utilize global image priors across the entire image space. To further enhance the performance, Cui \textit{et al.} \cite{9635679} propose a novel non-local neural network based CS framework (NL-CSNet), which captures long-range dependencies through an attention mechanism. In recent years, Transformer-based CS methods \cite{shen2024mtc}, \cite{9934025}, \cite{9841016} have been proposed to enhance image reconstruction quality. For example, TCS-Net \cite{10049603} introduces a Transformer-based deep reconstruction module that progressively refines the reconstruction by capturing global dependencies and contextual information across the entire image. Similarly, CSformer \cite{Ye2023CSformer} introduces a progressive reconstruction strategy, leveraging the Transformer's global context capture capabilities to incrementally restore image details. Apparently, compared to earlier block-by-block algorithms, the above methods inferenced in the global image space significantly alleviate block artifacts and obtain superior performance.

\subsection{Deep Unfolding Image CS Methods}

% Deep unfolding image CS methods unfold certain classical optimization algorithms into deep neural networks and usually provide a strong theoretical support and a powerful interpretability. Based on this idea, ISTA-Net \cite{zhang2018ista} attempts to transform the Iterative Shrinkage-Thresholding (ISTA) algorithm into a deep unfolding network, in which a neural network is utilized to approximate the proximal mapping related to the sparsity-inducing regularizer. Another typical example is AMP-Net \cite{9298950}, which simulates the iterative process of the Approximate Message Passing (AMP) algorithm and introduces a deep deblocking module. By unfolding traditional optimization algorithms, these networks \cite{zhang2022ams}, \cite{9854112}, \cite{2022FSOINet} have achieved better performance. However, these methods rely on transmitting reconstructed images across different stages, which hinders their ability to capture the complex dependencies within the images.

Deep unfolding methods aim to translate classical optimization algorithms into deep networks, thereby offering strong theoretical interpretability. 
For example, ISTA-Net~\cite{zhangISTANetInterpretableOptimizationInspired2018} unfolds the Iterative Shrinkage-Thresholding Algorithm (ISTA) and uses neural networks to approximate the proximal operator for sparsity. 
Similarly, AMP-Net~\cite{zhangAMPNetDenoisingBasedDeep2021} simulates the Approximate Message Passing (AMP) process and incorporates a learned deblocking module. 
However, these early unfolding schemes~\cite{zhangAMSNetAdaptiveMultiScale2022,chenContentAwareScalableDeep2022,chenFSOINETFeatureSpaceOptimizationInspired2022} typically exchange only the reconstructed images between stages, which inevitably leads to information loss and generally limits the final CS reconstruction quality. 

\begin{figure}[b]
\vskip -0.17in
  \centering
  % \fbox{\rule{0pt}{2in} \rule{0.95\linewidth}{0pt}}
  \includegraphics[width=1.0\linewidth]{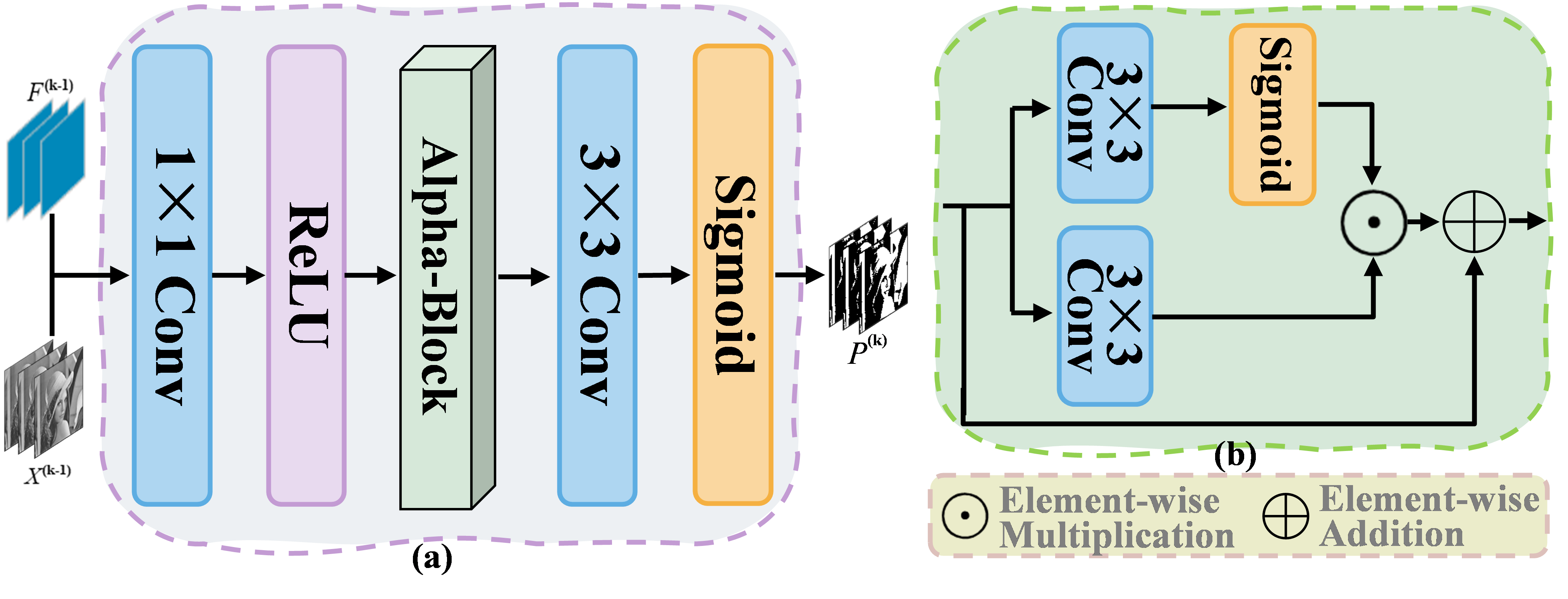}

\vskip -0.13in
   \caption{(a) shows the network architecture of our proposed AlphaNet, and (b) is the structure details of Alpha-Block.}
   \label{fig:3}
% \vskip -0.02in
\end{figure}

To overcome aforementioned  limitations, recent deep unfolding CS methods~\cite{wangUFCNetUnrollingFixedpoint2024,songDynamicPathControllableDeep2023,guoCPPNetEmbracingMultiScale2024} have begun to transmit both deep features and reconstructions across cascaded stages. 
For instance, SODAS-Net~\cite{songSODASnetSideinformationaidedDeep2023} extends ISTA-Net by introducing side information and exchanging richer representations between stages to boost representational capacity. 
Likewise, DUN-CSNet~\cite{cuiDeepUnfoldingNetwork2024} integrates intermediate features into the gradient descent and proximal mapping steps, leading to more effective modeling of complex image structures. 
Similar strategies are employed by OCTUF~\cite{songOptimizationInspiredCrossAttentionTransformer2023} and USB-Net~\cite{guoUSBNetUnfoldingSplit2025}, where networks iteratively pass and integrate intermediate features across stages to facilitate progressive refinement. 
By facilitating richer feature interactions across stages, these deep unfolding approaches achieve superior reconstruction performance.

\section{Proposed Method}
\label{sec:formatting}

\subsection{Preliminary Knowledge}
% \subsubsection{Multiplicity of Solutions in CS systems}

\textbf{Non-Uniqueness Property of CS:} In the context of image CS, assuming that \( \operatorname{rank}(\mathbf{A}) = r \), the rank-nullity theorem from linear algebra yields:
\begin{equation}
    \operatorname{rank}(\mathbf{A}) + \operatorname{nullity}(\mathbf{A}) = N,
\end{equation}
where \( \operatorname{nullity}(\mathbf{A}) = \dim(\operatorname{Null}(\mathbf{A})) \) denotes the dimension of the null space of \( \mathbf{A} \). Consequently, we have:
\begin{equation}
    \operatorname{nullity}(\mathbf{A}) = N - r \geq N - M > 0
\end{equation}
indicating that the null space of \( \mathbf{A} \) has positive dimension.

This implies that for any particular solution \( \mathbf{\hat{x}} \) satisfying \( \mathbf{A} \mathbf{\hat{x}} = \mathbf{y} \), the solution set forms an affine subspace:
\begin{equation}
    \mathcal{S} = \left\{ \mathbf{x} \in \mathbb{R}^N \mid \mathbf{x} = \mathbf{\hat{x}} + \mathbf{z}, \ \text{with} \quad \mathbf{z} \in \operatorname{Null}(\mathbf{A}) \right\}
\end{equation}
This subspace has a dimension equal to \( N - r \), reflecting the fact that the underdetermined nature of the measurement matrix leads to multiple mathematically valid solutions.

% As above, due to the inherent non-injectivity of \( \mathbf{A} \), the solution to the image CS problem is not unique but admits an entire affine subspace of candidates. Additional priors or regularization constraints are required to favor particular solutions within this null space.

% \subsubsection{Proximal Gradient Descent Algorithm}
\hspace{-0.22in} \textbf{Proximal Gradient Descent:} Proximal Gradient Descent (PGD) algorithm, as a prevailing optimization method, has been widely used to solve many large-scale linear inverse problems. Mathematically, the PGD algorithm solves Eq.~\eqref{e1} through the following iterative steps:
\vspace{-0.03in}
\begin{numcases}{}
\mathbf{r}^{(k)} = \mathbf{x}^{(k-1)} - \rho\mathbf{A}^{\rm T}(\mathbf{A} \mathbf{x}^{(k-1)}-\mathbf{y}) \label{eq22} \\
% \vspace{0.1in}
\mathbf{x}^{(k)} = \mathcal{\rm{prox}}_{\lambda}(\mathbf{r}^{(k)}) \label{eq33}
% \label{eq:6}
\end{numcases}
\vspace{-0.03in}
where Eq.~\eqref{eq22} is responsible for the gradient descent and $\rho$ is the pre-defined step size. In Eq.~\eqref{eq33}, ${\rm prox}_{\lambda}(\cdot)$ indicates a specific proximal operator. Corresponding to Eqs.~\eqref{eq22}~\eqref{eq33}, the existing PGD-inspired DUNs usually integrate deep networks with the traditional PGD algorithm to solve CS problem by iterating the following steps:  %  for controlling the intensity of gradient updating
\vspace{-0.06in}
\begin{numcases}{}
\textbf{r}^{(k)} = \textbf{x}^{(k-1)} - \rho^{(k)}\mathbf{A}^{\rm T}(\mathbf{A} \textbf{x}^{(k-1)}-\textbf{y}) \label{eq44} \\
% \vspace{0.1in}
\textbf{x}^{(k)} = \mathcal{H}^{(k)}_{r}(\textbf{r}^{(k)}) \label{eq55}
% \label{eq:1}
%\vspace{-0.01in}
\end{numcases}
where Eq.~\eqref{eq44} indicates the gradient descent of $k$-$th$ stage, and the parameter $\rho^{(k)}$ is the corresponding step size. Eq.~\eqref{eq55} corresponds to a specific proximal operator, which is usually fitted by a well-designed neural network ($\mathcal{H}^{(k)}_{r}$) to learn an efficient deep proximal mapping.

\subsection{Overall Architecture}

% Due to the inherent underdetermined nature of CS problems, multiple feasible solutions typically exist for a given set of observed measurements. However, the traditional approaches tend to recover a single estimate, potentially neglecting the diversity of solutions consistent with the observed data. As above, we propose a novel multi-hypothesis collaborative deep unfolding framework that jointly reconstructs and leverages multiple reconstructions to enhance overall performance.

The CS problem is inherently underdetermined, theoretically admitting multiple feasible solutions for a given measurement set. As above, we propose a novel multi-hypothesis collaborative deep unfolding framework that jointly reconstructs and integrates multiple hypotheses.

Specifically, our approach leverages the inherent ambiguity of CS by jointly reconstructing a set of hypotheses $\tilde{\mathbf{X}}=\{x_{1}, x_{2},...,x_{T}\}$. The optimization is represented by:
\vskip -0.2in
\begin{numcases}{}
\tilde{\mathbf{X}} = \arg \underset{\mathbf{X}}{\min} \frac{1}{2} \|\mathbf{A} \mathbf{X} - \mathbf{y} \|_2^2 + \lambda \mathbf{\Psi}(\mathbf{X}) \label{e6}\\
% \vspace{0.1in}
\tilde{\mathbf{x}} = \rm{Merg}(\tilde{\mathbf{X}}) \label{e7}
% \label{eq:6}
\end{numcases}
where $\mathbf{\Psi}(\mathbf{X})$ incorporates both intra-hypothesis priors and inter-hypothesis correlation priors, while Merg($\cdot$) aggregates multiple hypotheses into the final reconstructed image. To optimize this multi-hypothesis optimization model, we unfold its proximal gradient descent as:
\vskip -0.1in
\begin{empheq}[left=\empheqlbrace]{align}
		&\mathbf{R}^{\left(k\right)}=\mathbf{X}^{\left(k-1\right)}-\mathbf{P}^{\left(k\right)}\mathrm{\nabla \mathit{f}}\left(\mathbf{X}^{\left(k-1\right)}\right) \label{eq11} \\
		&\mathbf{X}^{\left(k+1\right)}=\mathcal{H}_{R}^{\left(k\right)}\left(\mathbf{R}^{\left(k\right)}\right)
	    \label{eq12}
\end{empheq}
where $\mathrm{\nabla \mathit{f}}\left(\cdot \right)$ is the gradient of the data fidelity term with respect to the hypotheses $\mathbf{X}^{\left(k\right)}$, and $\mathbf{P}^{\left(k\right)}$ contains step sizes for all hypotheses, and $\mathcal{H}_{R}^{\left(k\right)}$($\cdot$) is a deep network that learns the proximal mapping, incorporating both intra-hypothesis priors and inter-hypothesis correlations.

\begin{figure}[t]
% \vskip -0.16in
  \centering
  % \fbox{\rule{0pt}{2in} \rule{0.95\linewidth}{0pt}}
  \includegraphics[width=1.0\linewidth]{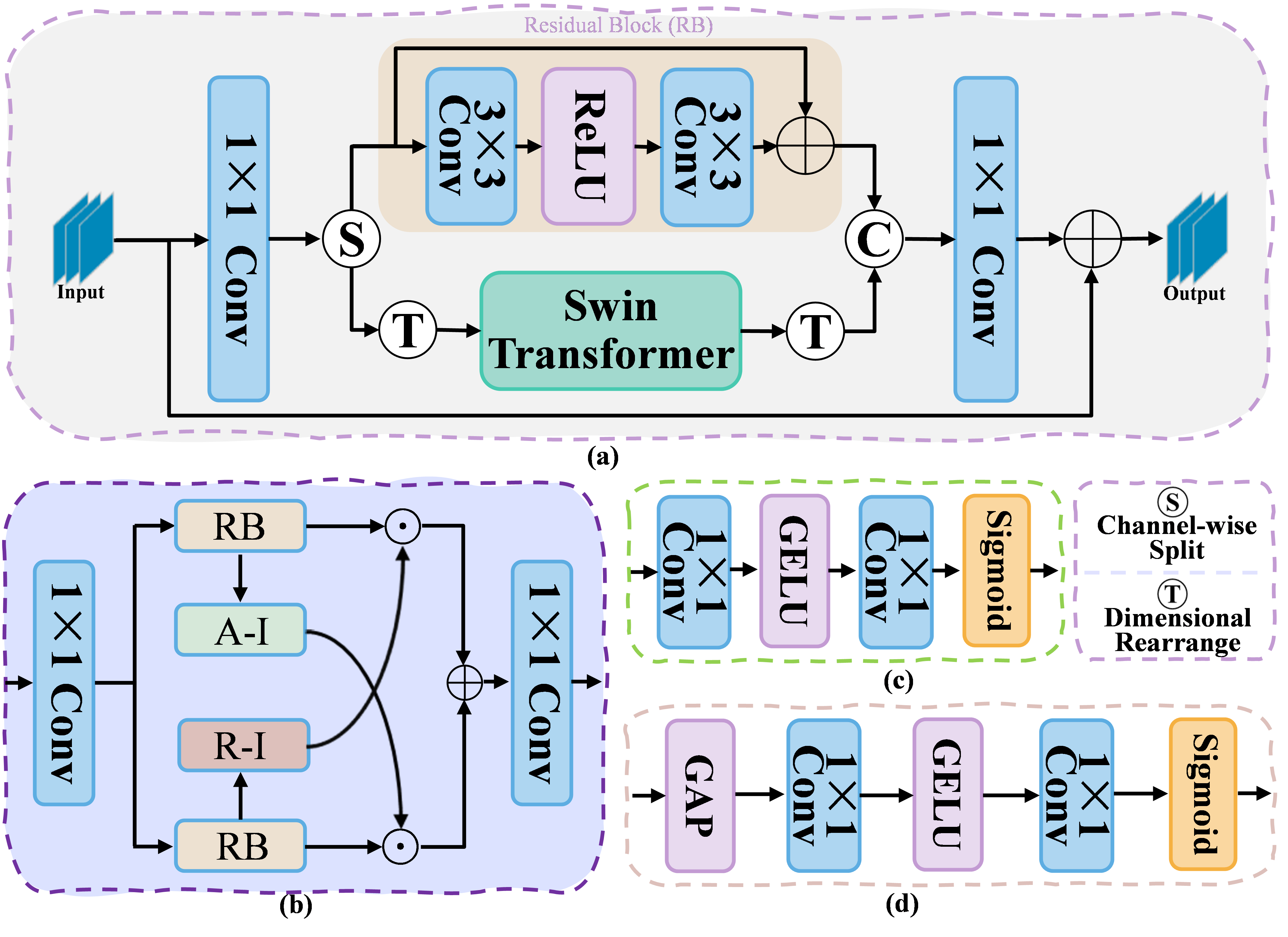}

\vskip -0.1in
   \caption{(a) is the architecture of FEB, (b) shows the details of block MHCB, and (c), (d) respectively indicate A-I and R-I blocks.}
   \label{fig:4}
\vskip -0.12in
\end{figure}

\begin{figure*}[t]
\begin{center}

\hspace{2.08in}
\includegraphics[width=0.9in]{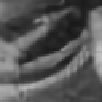}  \hspace{-0.004in}
\includegraphics[width=0.9in]{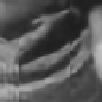}  \hspace{-0.004in}
\includegraphics[width=0.9in]{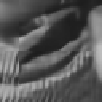}  \hspace{-0.004in}
\includegraphics[width=0.9in]{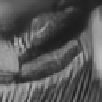}  \hspace{-0.004in}
\includegraphics[width=0.9in]{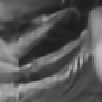} % \hspace{-0.004in}
\\
\vskip -0.01in \hskip 2.07in \scriptsize{ TransCS~\cite{shenTransCSTransformerBasedHybrid2022}\quad\quad \quad\quad\quad TCS-Net~\cite{ganPatchPixelTransformerBased2023} \quad\quad\quad\quad CSformer~\cite{yeCSformerBridgingConvolution2023} \quad\quad\quad \ \quad DPC-DUN~\cite{songDynamicPathControllableDeep2023} \quad\quad\quad\ \quad OCTUF~\cite{songOptimizationInspiredCrossAttentionTransformer2023}}
\vskip -0.02in \hskip 2.12in \scriptsize{(24.92/0.7672) \quad\quad\quad\ \quad (24.70/0.7450) \quad\quad\ \ \quad (26.62/0.8061) \quad\quad\quad\quad (26.43/0.8161) \quad\quad\quad\quad (25.13/0.7779)}
\vskip 0.02in
% \vspace{0.16in}
\hspace{2.088in}
\includegraphics[width=0.9in]{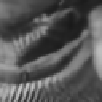}  \hspace{0.02in}
\includegraphics[width=0.9in]{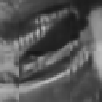}  \hspace{0.02in}
\includegraphics[width=0.9in]{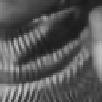}  \hspace{0.02in}
\includegraphics[width=0.9in]{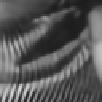}  \hspace{0.02in}
\includegraphics[width=0.9in]{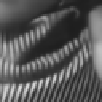} % \hspace{0.2in}
\\
\vskip -0.01in \hskip 1.93in \scriptsize{NesTD-Net~\cite{ganNesTDnetDeepNESTAinspired2024} \quad\quad\quad \quad UFC-Net~\cite{wangUFCNetUnrollingFixedpoint2024} \quad\quad\quad\quad CPP-Net~\cite{guoCPPNetEmbracingMultiScale2024} \quad\quad \quad\quad USB-Net~\cite{guoUSBNetUnfoldingSplit2025} \quad\quad\ \quad\quad \quad \textbf{OURS}}
\vskip -0.02in \hskip 2.1in \scriptsize{(26.59/0.8279) \quad\quad\quad\ \quad (25.40/0.7805) \quad\quad\ \ \ \quad (27.34/0.8395) \quad\quad\quad\quad (27.57/0.8558) \quad\quad\quad\ \textbf{(\color{cvprred}{29.37/0.8807})}}
\vskip 0.02in
\vspace{-2.286in}
\hspace{-4.86in}
\includegraphics[width=2.044in]{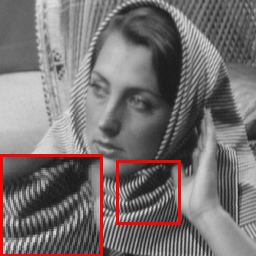}  %\hspace{0.05in}

\vskip -0.02in \hskip -4.72in \scriptsize{Ground Truth}
\vskip -0.01in \hskip -4.71in \scriptsize{(PSNR/SSIM)}

\end{center}
\vskip -0.2in
\caption{The visual comparisons between different CS methods for recovering an image from Set11 in the case of CS rate = 0.10.}
\vskip -0.07in
\label{fig:5}
\end{figure*}

% Placeholder tables
\begin{table*}[t]
\renewcommand\arraystretch{0.55}
\centering
\caption{Average PSNR/SSIM comparisons between different CS networks at various sampling rates on dataset Set11. \textcolor{cvprred}{Red} indicates the best result, \textcolor{cvprblue2}{blue} denotes the second best result, and \underline{underline} signifies the best performance of existing CS reconstruction networks.}
\label{tab:1}
\vspace{-0.12in}
\small

\begin{tabular}{p{3.62cm}<{\centering} | p{1.88cm}<{\centering} p{1.78cm}<{\centering} p{1.78cm}<{\centering} p{1.78cm}<{\centering} p{1.78cm}<{\centering} | p{1.78cm}<{\centering}}
\toprule
\multirow{2}*{Algorithms} & \multicolumn{5}{c|}{Sampling Ratio} & \multirow{2}*{Average}\\
\cline{2-6}
&\vspace{-0.04in}0.01\vspace{-0.03in}&\vspace{-0.04in}0.10\vspace{-0.03in}&\vspace{-0.04in}0.25\vspace{-0.03in}&\vspace{-0.04in}0.30\vspace{-0.03in}&\vspace{-0.04in}0.40\vspace{-0.03in}&\\
\midrule

\footnotesize{CSNet}\tiny{$_{\rm \textcolor{red}{(ICME2017)}}$}\footnotesize{~\cite{shiDeepNetworksCompressed2017}}&21.01/0.5560&28.10/0.8514&32.10/0.9221&33.86/0.9448&35.88/0.9605&30.19/0.8470\\
\footnotesize{LapCSNet}\tiny{$_{\rm \textcolor{red}{(ICASSP2018)}}$}\footnotesize{~\cite{cuiEfficientDeepConvolutional2018}}&21.54/0.5659&28.34/0.8571&--\ --/--\ --&--\ --/--\ --&--\ --/--\ --&--\ --/--\ --\\
\footnotesize{SCSNet}\tiny{$_{\rm \textcolor{red}{(CVPR2019)}}$}\footnotesize{~\cite{shiScalableConvolutionalNeural2019}}&21.04/0.5562&28.52/0.8616&33.43/0.9373&34.64/0.9511&36.92/0.9666&30.91/0.8546\\
\footnotesize{CSNet$^{+}$}\tiny{$_{\rm \textcolor{red}{(TIP2020)}}$}\footnotesize{~\cite{shiImageCompressedSensing2020}}&21.03/0.5566&28.34/0.8580&33.34/0.9387&34.27/0.9492&36.44/0.9690&30.68/0.8543\\
\footnotesize{DPA-Net}\tiny{$_{\rm \textcolor{red}{(TIP2020)}}$}\footnotesize{~\cite{sunDualPathAttentionNetwork2020}}&18.20/0.5101&27.66/0.8530&32.38/0.9311&33.35/0.9425&35.21/0.9580&29.36/0.8389\\
\footnotesize{TCS-Net}\tiny{$_{\rm \textcolor{red}{(TCI2023)}}$}\footnotesize{~\cite{ganPatchPixelTransformerBased2023}}&21.09/0.5505&29.04/0.8834&33.94/0.9508&35.49/0.9602&37.93/0.9750&31.50/0.8640\\
\footnotesize{NL-CSNet}\tiny{$_{\rm \textcolor{red}{(TMM2023)}}$}\footnotesize{~\cite{cuiImageCompressedSensing2023}}&21.96/0.6005&30.05/0.8995&34.45/0.9513&35.68/0.9606&37.71/0.9753&31.97/0.8774\\
\footnotesize{CSformer}\tiny{$_{\rm \textcolor{red}{(TIP2023)}}$}\footnotesize{~\cite{yeCSformerBridgingConvolution2023}}&21.63/0.5905&29.21/0.8784&33.36/0.9490&35.06/0.9523&37.20/0.9679&31.29/0.8676\\

\midrule

\footnotesize{FSIONet}\tiny{$_{\rm \textcolor{red}{(ICASSP2022)}}$}\footnotesize{~\cite{chenFSOINETFeatureSpaceOptimizationInspired2022}}         & 21.73/0.5937                                                             & 30.44/0.9018                                                                 & 35.80/0.9595                                                                 & 37.00/0.9665                                                                 & 39.13/0.9764                                                                 & 32.82/0.8796                                                                 \\
\footnotesize{CASNet}\tiny{$_{\rm \textcolor{red}{(TIP2022)}}$}\footnotesize{~\cite{chenContentAwareScalableDeep2022}}                            & 21.99/0.6085     & 30.35/0.9012                                                                 & 35.69/0.9593                                                                 & 36.93/0.9664                                                                 & 39.04/0.9761                                                                 & 32.80/0.8817                                                     \\
		\footnotesize{TransCS}\tiny{$_{\rm \textcolor{red}{(TIP2022)}}$}\footnotesize{~\cite{shenTransCSTransformerBasedHybrid2022}}                      & 20.68/0.5680                                                             & 29.54/0.8877                                                                 & 35.06/0.9548                                                                 & 35.62/0.9588                                                                 & 38.46/0.9737                                                                 & 31.87/0.8686                                                                 \\
		\footnotesize{AMS-Net}\tiny{$_{\rm \textcolor{red}{(TMM2022)}}$}\footnotesize{~\cite{zhangAMSNetAdaptiveMultiScale2022}}                          & 21.45/0.5795                                                             & 30.73/0.8867                                                     & 35.85/0.9522                                                                 & 36.97/0.9599                                                                 & 38.87/0.9702                                                                 & 32.77/0.8697                                                                 \\
		\footnotesize{DPC-DUN}\tiny{$_{\rm \textcolor{red}{(TIP2023)}}$}\footnotesize{~\cite{songDynamicPathControllableDeep2023}}                        & 20.54/0.5645                                                             & 29.40/0.8798                                                                 & 34.69/0.9482                                                                 & 35.88/0.9570                                                                 & 37.98/0.9694                                                                 & 31.70/0.8638                                                                 \\
		\footnotesize{OCTUF$^{+}$}\tiny{$_{\rm \textcolor{red}{(CVPR2023)}}$}\footnotesize{~\cite{songOptimizationInspiredCrossAttentionTransformer2023}} & 21.78/0.5941                                                             & 30.70/0.9030                                                     & 36.10/0.9604                                         & 37.21/0.9673                                         & 39.41/0.9773                                         & 33.04/0.8804                                                     \\
		\footnotesize{NesTD-Net}\tiny{$_{\rm \textcolor{red}{(TIP2024)}}$}\footnotesize{~\cite{ganNesTDnetDeepNESTAinspired2024}}                        & 21.40/0.5891                                                             & 30.91/0.9099                                                                 & 36.27/0.9622                                                                 & 37.17/0.9677                                                                 & 39.47/0.9778                                                                 & 33.04/0.8813                                                                 \\
		\footnotesize{CPP-Net}\tiny{$_{\rm \textcolor{red}{(CVPR2024)}}$}\footnotesize{~\cite{guoCPPNetEmbracingMultiScale2024}}                          & 22.19/0.6135                                                             & 31.27/0.9135                                                                 & 36.35/0.9631                                                                 & 37.55/0.9696                                                                 & 39.53/0.9781                                                                 & 33.38/0.8876                                                                 \\
		\footnotesize{USB-Net}\tiny{$_{\rm \textcolor{red}{(TIP2025)}}$}\footnotesize{~\cite{guoUSBNetUnfoldingSplit2025}}                               & \underline{22.29}/\underline{0.6168}                                                             & \underline{31.31}/\underline{0.9149}                                                                 & \underline{36.42}/\underline{0.9632}                                                                 & \underline{37.65}/\underline{0.9699}                                                                 & \underline{39.64}/\underline{0.9785}                                                                 & \underline{33.46}/\underline{0.8887}                                                                 \\

\midrule

MHC-DUN$^{\textbf{*}}$&\textbf{\textcolor{cvprblue2}{22.55}}/\textbf{\textcolor{cvprblue2}{0.6387}}&\textbf{\textcolor{cvprblue2}{31.82}}/\textbf{\textcolor{cvprblue2}{0.9206}}&\textbf{\textcolor{cvprblue2}{36.81}}/\textbf{\textcolor{cvprblue2}{0.9646}}&\textbf{\textcolor{cvprblue2}{38.05}}/\textbf{\textcolor{cvprblue2}{0.9708}}&\textbf{\textcolor{cvprblue2}{40.04}}/\textbf{\textcolor{cvprblue2}{0.9793}}&\textbf{\textcolor{cvprblue2}{33.85}}/\textbf{\textcolor{cvprblue2}{0.8947}}\\
MHC-DUN&\textbf{\textcolor{cvprred}{22.63}}/\textbf{\textcolor{cvprred}{0.6392}}&\textbf{\textcolor{cvprred}{31.86}}/\textbf{\textcolor{cvprred}{0.9208}}&\textbf{\textcolor{cvprred}{36.87}}/\textbf{\textcolor{cvprred}{0.9649}}&\textbf{\textcolor{cvprred}{38.10}}/\textbf{\textcolor{cvprred}{0.9710}}&\textbf{\textcolor{cvprred}{40.08}}/\textbf{\textcolor{cvprred}{0.9796}}&\textbf{\textcolor{cvprred}{33.91}}/\textbf{\textcolor{cvprred}{0.8951}}\\
\bottomrule

\end{tabular}
\vspace{-0.16in}
\end{table*}

% Based on this unfolded optimization, our framework introduces two collaboratively designed modules: the Multi-Hypothesis Collaborative Gradient Descent Module (MHC-GDM) and the Multi-Hypothesis Collaborative Proximal Mapping Module (MHC-PMM). Specifically, MHC-GDM is responsible for the joint gradient descent step across multiple hypotheses, utilizing the shared measurement consistency to update the entire solution set simultaneously. MHC-PMM, in turn, performs proximal mapping by exploiting both the internal prior within each candidate solution and the relational priors across multiple hypotheses, enabling richer representation and effective fusion of complementary information.

Aligned with Eqs.\eqref{eq11}\eqref{eq12}, our framework (Fig.~\ref{fig:2}) comprises two functional modules: Multi-Hypothesis Collaborative Gradient Descent Module (MHC-GDM) for joint gradient updates across hypotheses, and Multi-Hypothesis Collaborative Proximal Mapping Module (MHC-PMM), which leverages both intra and inter-hypothesis priors.

%-------------------------------------------------------------------------
\subsection{MHC-GDM for Gradient Descent}

% The MHC-GDM module is designed to perform a collaborative gradient update across multiple hypotheses. Unlike conventional methods, we introduce a learnable network, AlphaNet, to cooperatively predict spatially varying step size maps for all hypothesis, enabling dynamic and fine-grained step size control for each candidate reconstruction.

The MHC-GDM module (Fig.~\ref{fig:2}) is designed to perform collaborative gradient updates across multiple hypotheses. Instead of fixed steps, we introduce AlphaNet, a well-designed learnable network that cooperatively predicts spatially varying step size maps for all hypotheses, enabling dynamic and fine-grained refinement.

Formally, at $k$-$th$ iteration, the corresponding step size maps $\mathbf{P}^{\left(k\right)}$ for all hypotheses can be expressed as:
\vspace{-0.05in}
\begin{equation}
    \mathbf{P}^{\left(k\right)} = {\rm AlphaNet} (\mathbf{F}^{(k-1)}, \mathbf{X}^{(k-1)})
\vspace{-0.05in}
\end{equation}
where AlphaNet($\cdot$) indicates our proposed network, and it takes as input both the reconstructed hypotheses $\mathbf{X}^{(k-1)}$ and the features $\mathbf{F}^{(k-1)}$ from the previous stage.

Specifically, in the proposed AlphaNet (as shown in Fig.~\ref{fig:3}), an initial 1×1 convolution followed by a ReLU is employed to integrate the dual-domain information, i.e., image and feature domains, in a channel-wise manner:
\vspace{-0.07in}
\begin{equation}
\textbf{u}^{(k)} = {\rm ReLU}({\rm Conv}({\rm Cat}(\mathbf{F}^{(k-1)}, \mathbf{X}^{(k-1)})))
\vspace{-0.05in}
\label{jiji}
\end{equation}
where ${ \rm Cat}(\cdot)$ is channel-wise concatenation, enabling joint processing of these dual-domain heterogeneous data.

Subsequently, a well-designed Alpha-Block employs a residual spatial attention mechanism to refine fine-grained features by adaptively recalibrating spatial responses, facilitating collaborative learning across multiple hypotheses. Specifically, the proposed Alpha-Block consists of two parallel 3×3 convolutions, with one branch producing an attention map via a Sigmoid activation layer. Finally, the attention-weighted features are merged with the identity shortcut through element-wise addition:
\vspace{-0.07in}
\begin{equation}
\textbf{v}^{(k)} = \textbf{u}^{(k)} + {\rm Conv}(\textbf{u}^{(k)})\odot {\rm Sigmoid}({\rm Conv}(\textbf{u}^{(k)}))
\vspace{-0.05in}
\label{jiji}
\end{equation}
where $\odot$ signifies element-wise multiplication, facilitating attention-weighted feature modulation that adaptively accentuates informative interactions.

% Finally, a 3×3 convolution and Sigmoid operator normalizes the attention-weighted features onto the output step size maps:
Finally, a 3×3 convolution followed by Sigmoid normalizes the attention-weighted features into step size maps:
\vspace{-0.07in}
\begin{equation}
\textbf{P}^{(k)} = {\rm Sigmoid}({\rm Conv}(\textbf{v}^{(k)}))
\vspace{-0.05in}
\label{jiji}
\end{equation}

% Through this design, the proposed AlphaNet dynamically and jointly models spatially varying step sizes for multiple hypotheses, effectively bridging dual-domain cues and leveraging fine-grained attention to improve reconstruction quality and convergence stability in the multi-hypothesis collaborative framework.

Through this design, the proposed AlphaNet is able to dynamically generate spatially adaptive step size maps for all hypotheses, thereby realizing dynamic, fine-grained and hypothesis-specific gradient updating, thereby enhancing CS reconstruction performance.

%-------------------------------------------------------------------------
\subsection{ MHC-PMM for Proximal Mapping}

The module MHC-PMM is designed to collaboratively refine multiple hypotheses through a proximal mapping network that effectively performs joint denoising.

In the proposed MHC-PMM (Fig.~\ref{fig:2}), the dual-domain information from the image and deep feature domains is aggregated via concatenation and convolution:
\vspace{-0.07in}
\begin{equation}
\textbf{q}^{(k)} = {\rm Conv}({\rm Cat}({\rm Conv}( \mathbf{R}^{(k)}), \mathbf{F}^{(k-1)}))
\vspace{-0.05in}
\label{jiji}
\end{equation}
where ${\rm Cat}(\cdot)$ denotes channel-wise concatenation.

\begin{figure*}[t]
  \centering
  % \fbox{\rule{0pt}{2in} \rule{0.95\linewidth}{0pt}}
  \includegraphics[width=0.98\linewidth]{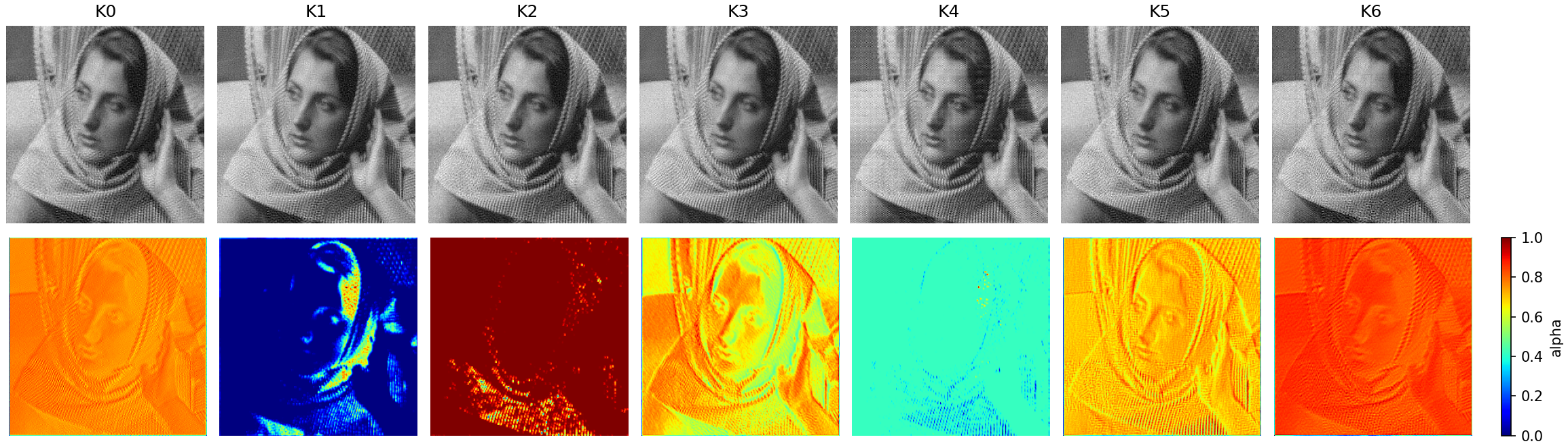}

\vskip -0.09in
   \caption{The first row depicts multiple hypotheses obtained within the $5$-$th$ stage, while the second row shows the corresponding step size maps learned by AlphaNet. This figure clearly demonstrates inter-hypothesis diversity and distinct step sizes for adaptive gradient updates.}
   \label{fig:6}
\vskip -0.18in
\end{figure*}

\begin{figure}[b]
\vskip -0.16in
  \centering
  % \fbox{\rule{0pt}{2in} \rule{0.95\linewidth}{0pt}}
  \includegraphics[width=1.0\linewidth]{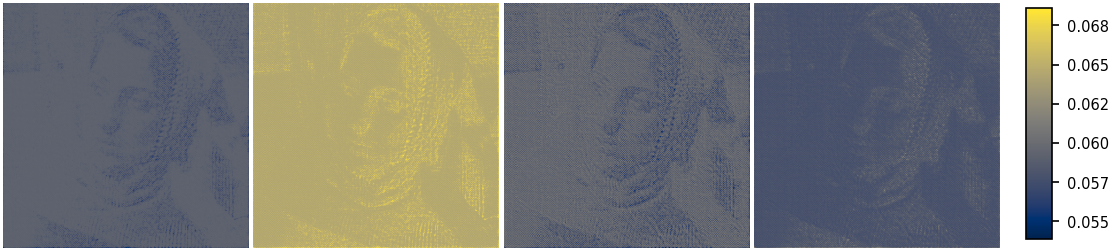}
  \vskip -0.05in \small{(a)}\quad \quad \quad\quad \quad \small{(b)}\quad \quad \quad\quad \quad \small{(c)}\quad  \quad\quad\quad \quad \small{(d)}\quad\quad \quad

\vskip -0.08in
   \caption{The visualization of certain spatial attention maps learned within the MHCB block.}
   \label{fig:7}
% \vskip -0.2in
\end{figure}

Subsequently, $d$ Feature Extraction Blocks (FEBs) as shown in Fig.~\ref{fig:4}(a) are employed to enhance the features $\textbf{q}^{(k)}$. Specifically, each FEB consists of two parallel branches: one utilizes convolutional operations to capture local priors, while the other incorporates a Swin Transformer to model non-local dependencies. This dual-branch design effectively integrates local and global information, enhancing the robustness of feature representations.

% After FEBs, the enhanced feature $ \mathbf{F}^{(k)}$ is generated, followed which $T$ reconstruction heads $\{\mathcal{H}_{t}^{(k)}\}_{t=1}^{T}$ independently generate multiple hypotheses:: 
Following the FEBs, the enhanced feature  $\mathbf{F}^{(k)}$ is fed into $T$ reconstruction heads $\{\mathcal{H}_{t}^{(k)}\}_{t=1}^{T}$ to independently generate multiple hypotheses:
\vspace{-0.07in}
\begin{equation}
\{\mathbf{\hat{x}}_{1}^{(k)} \hdots \  \mathbf{\hat{x}}_{T}^{(k)} \} = \{\mathcal{H}_{1}^{(k)}(\mathbf{F}^{(k)}) \ \hdots \ \mathcal{H}_{T}^{(k)}(\mathbf{F}^{(k)}) \}
\vspace{-0.05in}
\label{jiji}
\end{equation}
where $\mathcal{H}_{i}^{(k)}$ is the $i$-$th$ reconstruction head, realized via a convolutional layer, that outputs the $i$-$th$ hypothesis $\mathbf{\hat{x}}_{i}^{(k)}$.

% Finally, a novel Multi-Hypothesis Collaborative Block (MHCB) is designed to explore correlations among the multiple hypotheses. Specifically, the proposed MHCB exploits channel and spatial attention mechanisms to perform coarse and fine-grained fusion of these hypotheses, capturing their individual and inter-correlations priors effectively:

Finally, a novel Multi-Hypothesis Collaborative Block (MHCB) as shown in Fig.~\ref{fig:4}(b) is designed to explore correlations among the multiple hypotheses. Specifically, the proposed MHCB leverages channel and spatial attention mechanisms for coarse and fine-grained fusion, effectively modeling their intra- and inter-hypothesis priors:
\vspace{-0.07in}
\begin{equation}
\textbf{X}^{(k)} = {\rm MHCB}({\rm Cat}(\mathbf{\hat{x}}_{1}^{(k)} \hdots \  \mathbf{\hat{x}}_{T}^{(k)}))
\vspace{-0.05in}
\label{jiji}
\end{equation}
where $\textbf{X}^{(k)}$ =$\{x_{1}^{(k)}\hdots x_{T}^{(k)}\}$, and ${\rm MHCB}(\cdot)$ is our MHCB.

Overall, the proposed MHC-PMM module effectively leverages comprehensive local and global priors through the combined use of CNN and Transformer architectures, and further facilitates collaborative refinement of multiple hypotheses by learning their inherent inter-correlations.

\subsection{Loss Function}

To enable end-to-end training, we design a composite loss applied at each unfolded stage, consisting of: \textbf{1)} a data fidelity term ($\mathcal{L}_{data}$) enforcing measurement consistency for each hypothesis; \textbf{2)} a diversity regularizer ($\mathcal{L}_{div}$) promoting complementarity among hypothesis. A final recovery loss ($\mathcal{L}_{rec}$) aligns the fused output with the ground truth:
\vspace{-0.06in}
\begin{equation}
\mathcal{L}({\mathbf{\Theta}}) =  \frac{1}{K} \sum_{k=1}^{K} (\lambda_{1} \mathcal{L}_{data}^{(k)} + \lambda_{2} \mathcal{L}_{div}^{(k)}) + \mathcal{L}_{rec}
\vspace{-0.05in}
\label{hahaha}
\end{equation}
where $k$ denotes the stage index, $\lambda_{1}$ and $\lambda_{2}$ are hyperparameters balancing the loss components, and $\mathbf{\Theta}$ represents the set of learnable parameters within the model.

The term $\mathcal{L}_{data}^{(k)}$ ensures that each hypothesis adheres to the observed measurements in the compressed domain:
\vspace{-0.07in}
\begin{equation}
\mathcal{L}_{data}^{(k)} =  \frac{1}{T} \sum_{i=1}^{T}\| A \mathbf{x}_{i}^{(k)} -  y \|_{2}^{2}
\vspace{-0.05in}
\label{hahaha}
\end{equation}
where $T$ is the number of hypotheses and $i$ is the index.

The term $\mathcal{L}_{div}^{(k)}$ penalizes similarity among hypotheses, promoting diversity and complementarity by minimizing their pairwise cosine similarity:
\vspace{-0.07in}
\begin{equation}
\mathcal{L}_{div}^{(k)} =  \frac{1}{T(T-1)} \sum_{i=1}^{T} \sum_{j\neq i}^{T} \frac{\langle \mathbf{x}_i^{(k)}, \mathbf{x}_j^{(k)}\rangle}{\|\mathbf{x}_i^{(k)}\|_2 \|\mathbf{x}_j^{(k)}\|_2}
\vspace{-0.05in}    
\label{hahaha}
\end{equation}
where $\langle \cdot \rangle$ represents the inner product operator.

Finally, the reconstruction loss $\mathcal{L}_{rec}$ ensures the quality of the final integrated solution:
\vspace{-0.07in}
\begin{equation}
\mathcal{L}_{rec} =  \| \tilde{\mathbf{x}} - \mathbf{x}  \|_{2}^{2}
\vspace{-0.05in}
\label{hahaha}
\end{equation}
where $\tilde{\mathbf{x}}$  is the fused output and  $\mathbf{x}$ is the ground truth image.

Apparently, the designed composite loss balances measurement fidelity, hypothesis diversity, and reconstruction accuracy, guiding the network to explore a rich solution space while ensuring consistent, high-quality recovery.

\begin{table*}[t]
\renewcommand\arraystretch{0.55}
\centering
\caption{Average PSNR/SSIM comparisons between different CS networks at various sampling rates on dataset Urban100. \textcolor{cvprred}{Red} indicates the best result, \textcolor{cvprblue2}{blue} denotes the second best result, and \underline{underline} signifies the best performance of existing CS reconstruction networks.}
\label{tab:2}
\vspace{-0.12in}
\small

\begin{tabular}{p{3.62cm}<{\centering} | p{1.88cm}<{\centering} p{1.78cm}<{\centering} p{1.78cm}<{\centering} p{1.78cm}<{\centering} p{1.78cm}<{\centering} | p{1.78cm}<{\centering}}
\toprule
\multirow{2}*{Algorithms} & \multicolumn{5}{c|}{Sampling Ratio} & \multirow{2}*{Average}\\
\cline{2-6}
&\vspace{-0.04in}0.01\vspace{-0.03in}&\vspace{-0.04in}0.10\vspace{-0.03in}&\vspace{-0.04in}0.25\vspace{-0.03in}&\vspace{-0.04in}0.30\vspace{-0.03in}&\vspace{-0.04in}0.40\vspace{-0.03in}&\\
\midrule

\footnotesize{CSNet}\tiny{$_{\rm \textcolor{red}{(ICME2017)}}$}\footnotesize{~\cite{shiDeepNetworksCompressed2017}}&20.74/0.5342&26.93/0.8006&30.98/0.9209&32.16/0.9316&33.80/0.9536&28.92/0.8282\\
\footnotesize{LapCSNet}\tiny{$_{\rm \textcolor{red}{(ICASSP2018)}}$}\footnotesize{~\cite{cuiEfficientDeepConvolutional2018}}&20.98/0.5456&27.25/0.8141&--\ --/--\ --&--\ --/--\ --&--\ --/--\ --&--\ --/--\ --\\
\footnotesize{SCSNet}\tiny{$_{\rm \textcolor{red}{(CVPR2019)}}$}\footnotesize{~\cite{shiScalableConvolutionalNeural2019}}&20.75/0.5343&27.06/0.8084&31.51/0.9247&32.97/0.9349&34.74/0.9584&29.41/0.8321\\
\footnotesize{CSNet$^{+}$}\tiny{$_{\rm \textcolor{red}{(TIP2020)}}$}\footnotesize{~\cite{shiImageCompressedSensing2020}}&20.76/0.5348&27.09/0.8086&31.44/0.9233&32.50/0.9341&34.29/0.9560&29.22/0.8314\\
\footnotesize{DPA-Net}\tiny{$_{\rm \textcolor{red}{(TIP2020)}}$}\footnotesize{~\cite{sunDualPathAttentionNetwork2020}}&17.97/0.4673&26.28/0.8084&30.60/0.9087&31.52/0.9236&33.50/0.9470&27.97/0.8110\\
\footnotesize{TCS-Net}\tiny{$_{\rm \textcolor{red}{(TCI2023)}}$}\footnotesize{~\cite{ganPatchPixelTransformerBased2023}}&20.93/0.5214&27.17/0.8400&31.40/0.9297&33.40/0.9391&35.76/0.9643&29.73/0.8389\\
\footnotesize{NL-CSNet}\tiny{$_{\rm \textcolor{red}{(TMM2023)}}$}\footnotesize{~\cite{cuiImageCompressedSensing2023}}&21.08/0.5514&28.43/0.8506&32.96/0.9410&33.62/0.9426&35.55/0.9641&30.33/0.8499\\
\footnotesize{CSformer}\tiny{$_{\rm \textcolor{red}{(TIP2023)}}$}\footnotesize{~\cite{yeCSformerBridgingConvolution2023}}&20.92/0.5482&27.45/0.8352&31.98/0.9282&33.18/0.9377&35.31/0.9614&29.77/0.8421\\

\midrule

\footnotesize{FSIONet}\tiny{$_{\rm \textcolor{red}{(ICASSP2022)}}$}\footnotesize{~\cite{chenFSOINETFeatureSpaceOptimizationInspired2022}}         & 20.97/0.5508                                                                 & 28.84/0.8712                                                                 & 33.84/0.9467                                                                 & 35.07/0.9571                                                                 & 37.12/0.9711                                                                 & 31.17/0.8594                                                                 \\
		\footnotesize{CASNet}\tiny{$_{\rm \textcolor{red}{(TIP2022)}}$}\footnotesize{~\cite{chenContentAwareScalableDeep2022}}                            & \underline{21.13}/\underline{0.5735}                                         & 28.78/0.8703                                                                 & 33.47/0.9440                                                                 & 34.62/0.9548                                                                 & 36.70/0.9694                                                                 & 30.94/0.8624                                                     \\
		\footnotesize{TransCS}\tiny{$_{\rm \textcolor{red}{(TIP2022)}}$}\footnotesize{~\cite{shenTransCSTransformerBasedHybrid2022}}                      & 20.86/0.5492                                                                 & 27.97/0.8495                                                                 & 32.90/0.9370                                                                 & 33.13/0.9418                                                                 & 36.43/0.9672                                                                 & 30.26/0.8489                                                                 \\
		\footnotesize{AMS-Net}\tiny{$_{\rm \textcolor{red}{(TMM2022)}}$}\footnotesize{~\cite{zhangAMSNetAdaptiveMultiScale2022}}                          & 20.94/0.5402                                                                 & \underline{29.09}/0.8519                                                     & 34.23/0.9383                                                     & 35.05/0.9499                                                     & 37.07/0.9650                                                     & 31.32/0.8491                                                     \\
		\footnotesize{DPC-DUN}\tiny{$_{\rm \textcolor{red}{(TIP2023)}}$}\footnotesize{~\cite{songDynamicPathControllableDeep2023}}                        & 20.43/0.5369                                                                 & 28.38/0.8502                                                                 & 33.69/0.9383                                                                 & 34.86/0.9499                                                                 & 36.92/0.9658                                                                 & 30.86/0.8482                                                                 \\
		\footnotesize{OCTUF$^{+}$}\tiny{$_{\rm \textcolor{red}{(CVPR2023)}}$}\footnotesize{~\cite{songOptimizationInspiredCrossAttentionTransformer2023}} & 21.03/0.5503                                                                 & 29.08/0.8715                                                    & \underline{34.27}/0.9487                                                     & \underline{35.46}/0.9587                                                     & \underline{37.51}/0.9721                                                     & \underline{31.47}/0.8603                                                                 \\
		\footnotesize{NesTD-Net}\tiny{$_{\rm \textcolor{red}{(TIP2024)}}$}\footnotesize{~\cite{ganNesTDnetDeepNESTAinspired2024}}                        & 20.13/0.5288                                                                 & 27.80/0.8681                                                                 & 33.02/0.9448                                                                 & 33.46/0.9515                                                                 & 35.90/0.9682                                                                 & 30.06/0.8576                                                                 \\
		\footnotesize{CPP-Net}\tiny{$_{\rm \textcolor{red}{(CVPR2024)}}$}\footnotesize{~\cite{guoCPPNetEmbracingMultiScale2024}}                          & 20.55/0.5554                                                                 & 28.49/0.8801                                                                 & 33.38/0.9485                                                                 & 34.58/0.9586                                                                 & 36.50/0.9714                                                                 & 30.70/0.8628                                                                 \\
		\footnotesize{USB-Net}\tiny{$_{\rm \textcolor{red}{(TIP2025)}}$}\footnotesize{~\cite{guoUSBNetUnfoldingSplit2025}}                               & 20.64/0.5617                                                                 & 28.58/\underline{0.8818}                                                                 & 33.56/\underline{0.9500}                                                                 & 34.79/\underline{0.9601}                                                                 & 36.72/\underline{0.9723}                                                                 & 30.86/\underline{0.8652}                                                                 \\

\midrule

MHC-DUN$^{\textbf{*}}$&\textbf{\textcolor{cvprblue2}{21.13}}/\textbf{\textcolor{cvprblue2}{0.5942}}&\textbf{\textcolor{cvprblue2}{29.92}}/\textbf{\textcolor{cvprblue2}{0.8935}}&\textbf{\textcolor{cvprblue2}{34.66}}/\textbf{\textcolor{cvprblue2}{0.9542}}&\textbf{\textcolor{cvprblue2}{35.81}}/\textbf{\textcolor{cvprblue2}{0.9625}}&\textbf{\textcolor{cvprblue2}{37.78}}/\textbf{\textcolor{cvprblue2}{0.9742}}&\textbf{\textcolor{cvprblue2}{31.86}}/\textbf{\textcolor{cvprblue2}{0.8757}}\\
MHC-DUN&\textbf{\textcolor{cvprred}{21.18}}/\textbf{\textcolor{cvprred}{0.5946}}&\textbf{\textcolor{cvprred}{29.97}}/\textbf{\textcolor{cvprred}{0.8938}}&\textbf{\textcolor{cvprred}{34.69}}/\textbf{\textcolor{cvprred}{0.9545}}&\textbf{\textcolor{cvprred}{35.87}}/\textbf{\textcolor{cvprred}{0.9627}}&\textbf{\textcolor{cvprred}{37.81}}/\textbf{\textcolor{cvprred}{0.9744}}&\textbf{\textcolor{cvprred}{31.90}}/\textbf{\textcolor{cvprred}{0.8760}}\\
\bottomrule

\end{tabular}
\vspace{-0.22in}
\end{table*}

%-------------------------------------------------------------------------
\section{Experiments and Analysis}

\subsection{Implementation Details}

In our MHC-DUN, a block-based sampling strategy with a block size of 32$\times$32 is employed. The network unfolds over $K$=10 stages, and within each stage, the number of FEBs is set to $d$=2 to balance the performance and computational cost. The number of hypotheses is set as $T$=16, enabling effective multi-hypothesis exploration. Besides, in our model MHC-DUN, the channel dimension of intermediate features is set to 128, and by sharing parameters across stages, we derive an alternative variant, denoted as MHC-DUN$^{\textbf{*}}$. In the composite loss function, the hyperparameters are empirically tuned as $\lambda_{1}$=0.50 and $\lambda_{2}$=0.01 to harmonize measurement fidelity, hypothesis diversity, and reconstruction accuracy.  \textit{For an analysis of the parameter configuration, please refer to the supplementary materials}.

% In the proposed MHC-DUN, the block size $B=32$. As depicted in Fig. \ref{fig:9}, the reconstructed quality converges with increasing phase number $K=15$ as performance saturates. For more configuration details, we set the channel number of the intermediate feature maps as 32, i.e., ${\rm c}=32$. For the number of FEB in sub-network SSG-Net, left figure of Fig.~\ref{fig:94} shows the relationship between the number of FEB and reconstructed quality, from which we can observe that with the increase of the number of FEB, the performance of the model gradually improves and tends to be stable. As above, in our proposed DUN-CSNet, the number of FEB in SSG-Net is set as 3. Similar to the analysis of FEB, right figure of Fig.~\ref{fig:94} shows the analysis results about the number of DINLM in DN-PMN, and finally we set the number of DINLM as 1.

\begin{table}[b]
\vskip -0.18in
\renewcommand\arraystretch{0.8}
\centering
  \caption{Performance comparison (PSNR/SSIM) of recent CS-MRI methods on the Brain dataset across different sampling ratios. (R indicates the sampling rate.)}
  \label{tab:3}
  \vspace{-0.12in}

  % \begin{tabular}{c|ccc|c}
  \begin{tabular}{p{2.0cm}<{\centering} | p{1.55cm}<{\centering}  p{1.55cm}<{\centering}  p{1.55cm}<{\centering}}
    \toprule

    \rowcolor{cvprgray} \small{Algorithms}&\small{R=0.10}&\small{R=0.20}&\small{Avg.}\\
    \midrule

        \small{DC-CNN~\cite{schlemper2017deep}} &\small{34.33/0.8957} &\small{38.43/0.9467} &\small{36.38/0.9212}\\
        \small{RDN~\cite{sun2018compressed}} &\small{34.38/0.8998}& \small{38.47/0.9474}& \small{36.43/0.9236}\\
        \small{ISTA-Net~\cite{zhangISTANetInterpretableOptimizationInspired2018}} &\small{34.62/0.9035} &\small{38.57/0.9478} &\small{36.60/0.9257}\\
        \small{CDNN~\cite{zheng2019cascaded}}& \small{34.67/0.9014} &\small{38.65/0.9476}& \small{36.66/0.9245}\\
        %ADMM-CSNet&31.37/0.8608 &34.45/0.8985& 38.52/0.9471 &34.78/0.9021\\
        \small{HiTDUN~\cite{zhang2022high}} &\small{35.71/0.9179}& \small{39.27/0.9529}& \small{37.49/0.9354}\\
        %CPP-Net~\cite{10655712} &34.87/0.9176& 36.61/0.9318 &39.67/0.9559 &37.05/0.9351\\
        \small{USB-Net~\cite{guoUSBNetUnfoldingSplit2025}}&\small{36.29/0.9275}&\small{39.52/0.9549}&\small{37.90/0.9412}\\

\midrule
    \small{MHC-DUN} & \small{\textbf{36.41/0.9279}} &\small{\textbf{39.65/0.9554}} & \small{\textbf{38.03/0.9417}}\\
   
  \bottomrule
\end{tabular}
% \vspace{-0.22in}
\label{tab:3}
\end{table}

For training, we use Waterloo Exploration Database (WED) as our training data. To diversify the training set, we apply random augmentations including 90$^{\circ}$, 180$^{\circ}$ and 270$^{\circ}$ rotations, as well as horizontal and vertical flips. During training, RGB images are converted to grayscale and randomly cropped to $128\times 128$ patches. We use PyTorch and train on an NVIDIA RTX 3090 GPU with the Adam optimizer ($\beta_{1}$=0.9 and $\beta_{2}$=0.999). Besides, we set the batch size as 16, and the initial learning rate is 1e-4, which is halved every 50 epochs. The model is trained for 600 epochs, each with 1000 iterations, totaling 600,000 iterations.

%-------------------------------------------------------------------------
\subsection{Comparisons with Other Methods}

We evaluate our method against two categories of compressive sensing (CS) algorithms: Deep Black Box Networks (DBNs) and Deep Unfolding Networks (DUNs). Specifically, we compare with eight DBNs (CSNet~\cite{shiDeepNetworksCompressed2017}, LapCSNet~\cite{cuiEfficientDeepConvolutional2018}, SCSNet~\cite{shiScalableConvolutionalNeural2019}, CSNet$^{+}$~\cite{shiImageCompressedSensing2020}, DPA-Net~\cite{sunDualPathAttentionNetwork2020}, TCS-Net~\cite{ganPatchPixelTransformerBased2023}, NL-CSNet~\cite{cuiImageCompressedSensing2023}, CSformer~\cite{yeCSformerBridgingConvolution2023}) and nine representative DUNs (FSIONet~\cite{chenFSOINETFeatureSpaceOptimizationInspired2022}, CASNet~\cite{chenContentAwareScalableDeep2022}, TransCS~\cite{shenTransCSTransformerBasedHybrid2022}, AMS-Net~\cite{zhangAMSNetAdaptiveMultiScale2022}, DPC-DUN~\cite{songDynamicPathControllableDeep2023}, OCTUF$^{+}$~\cite{songOptimizationInspiredCrossAttentionTransformer2023}, NesTD-Net~\cite{ganNesTDnetDeepNESTAinspired2024}, CPP-Net~\cite{guoCPPNetEmbracingMultiScale2024}, USB-Net~\cite{guoUSBNetUnfoldingSplit2025}). Experiments are conducted on Set11~\cite{kulkarniReconNetNonIterativeReconstruction2016} and Urban100~\cite{huangSingleImageSuperresolution2015} datasets across sampling rates of 0.01, 0.10, 0.25, 0.30, and 0.40.

%%%%%%%%%%%%%%%%%%Furthermore, the reconstructed results are evaluated with two commonly used assessment criteria: PSNR and SSIM at different sampling rates.

% Specifically, for the compared CS algorithms, when there is a pre-trained model at a given sampling rate, we directly fine tune the model using the same training data and augmentation policy. While when there is no pre-trained model, we directly train the model from scratch.

%
%\begin{figure}[b]
%\begin{center}
%\vskip -0.2in
%\includegraphics[width=1.58in]{visual4/20.png}  \hspace{-0.001in}
%\includegraphics[width=1.58in]{visual4/20.png}  \hspace{-0.001in}
%
%\end{center}
%\vskip -0.16in
%\caption{Multiple solutions and their weights and the fused version. under different learning rates (lr) for the testing image in Figure~\ref{fig:5}. (Patch size is 256$\times$256, Sampling rate is 0.10).}
%% \vskip -0.23in
%\label{fig:8}
%\end{figure}

\begin{table}[b]
\vskip -0.2in
\renewcommand\arraystretch{0.6}
\centering
  \caption{Complexity analysis (Parameters, GFLOPs and running time) of our proposed MHC-DUN and other competing methods on a 256$\times$256 input image at a CS ratio of 0.10.}
  \label{tab:4}
  \vspace{-0.12in}

  % \begin{tabular}{c|ccc|c}
  \begin{tabular}{p{2.0cm}<{\centering} | p{0.95cm}<{\centering}  p{0.95cm}<{\centering}  p{1.1cm}<{\centering}  p{1.1cm}<{\centering}}
    \toprule

    \rowcolor{cvprgray} \footnotesize{Algorithms}&\footnotesize{CPU/(s)}&\footnotesize{GPU/(s)}&\footnotesize{Param(M)}&\footnotesize{GFLOPs}\\
    \midrule
    \footnotesize{NesTD-Net~\cite{ganNesTDnetDeepNESTAinspired2024}} & \small{1.6431} &\small{0.1223} & \small{5.57}& \small{372.80}\\
    \footnotesize{CPP-Net~\cite{guoCPPNetEmbracingMultiScale2024}} & \small{1.3648} &\small{0.1182} & \small{12.31}& \small{166.93}\\
    \footnotesize{USB-Net~\cite{guoUSBNetUnfoldingSplit2025}} & \small{0.8825} &\small{0.0554} & \small{15.47}& \small{\textbf{95.89}}\\
\midrule
    \footnotesize{MHC-DUN$^{\textbf{*}}$} & \small{\textbf{0.8369}} &\small{\textbf{0.0512}} & \small{\textbf{3.68}}& \small{231.67}\\
    \footnotesize{MHC-DUN} & \small{0.9953} &\small{0.0627} & \small{10.81}& \small{231.67}\\
  \bottomrule
\end{tabular}
% \vspace{-0.22in}
\label{tab:4}
\end{table}

% NesTD-Net~\cite{ganNesTDnetDeepNESTAinspired2024}, CPP-Net~\cite{guoCPPNetEmbracingMultiScale2024}, USB-Net~\cite{guoUSBNetUnfoldingSplit2025}).

% \textbf{Comparisons with DBNs: }The experimental results on the given testing datasets are shown in Tables~\ref{tab:1} and~\ref{tab:2}, from which we can observe that the proposed network outperforms existing deep black box CS methods. Specifically, since the methods NL-CSNet~\cite{cuiImageCompressedSensing2023}, CSformer~\cite{yeCSformerBridgingConvolution2023} achieve the best performance, we mainly analyze the comparisons against these two CS algorithms. Specifically, \textbf{1)} On dataset Set11, the proposed MHC-DUN achieves on average 1.45dB, 2.13dB and 0.0110, 0.0208 gains in PSNR and SSIM compared against these two DBNs. \textbf{2)} On dataset Urban100, our proposed framework achieves on average 2.10dB, 2.66dB and 0.0248, 0.0326 gains in PSNR and SSIM. The visual comparisons are shown in Fig.~\ref{fig:5}, from which we observe that our proposed CS network can preserve more textural details compared with other DBNs.

\textbf{Comparisons with DBNs: }The experimental results on the testing datasets, summarized in Tables~\ref{tab:1} and \ref{tab:2}, demonstrate that our proposed network surpasses existing deep black box CS methods. Notably, we focus our comparison on the top-performing algorithms NL-CSNet~\cite{cuiImageCompressedSensing2023} and CSformer~\cite{yeCSformerBridgingConvolution2023}. Specifically, \textbf{1)} On the Set11 dataset, our MHC-DUN achieves average gains of 1.94 dB and 2.62 dB in PSNR, and 0.0177 and 0.0275 in SSIM, respectively. \textbf{2)} On the Urban100 dataset, the proposed framework attains improvements of 1.57 dB and 2.13 dB in PSNR, and 0.0261 and 0.0339 in SSIM. The visual comparisons in Fig.~\ref{fig:5} illustrate that our proposed CS approach better preserves textural details compared with other DBNs.

\textbf{Comparisons with DUNs: }Tables~\ref{tab:1} and~\ref{tab:2} present the experimental results across different datasets, demonstrating that our CS method consistently achieves superior reconstruction quality. Among the compared DUNs, recent methods CPP-Net~\cite{guoCPPNetEmbracingMultiScale2024} and USB-Net~\cite{guoUSBNetUnfoldingSplit2025} deliver the best performance. For clarity, we focus our analysis on these two CS algorithms. Specifically, \textbf{1)} On the Set11 dataset, our MHC-DUN achieves average gains of 0.53 dB and 0.45 dB in PSNR, 0.0075 and 0.0275 in SSIM, respectively. \textbf{2)} On the Urban100 dataset, the proposed framework attains improvements of 1.20 dB and 1.04 dB in PSNR, 0.0132 and 0.0108 in SSIM. The visual comparisons in Fig.~\ref{fig:5} indicate that our method more effectively preserves details and produces sharper edges compared to recent DUNs.

\begin{table}[t]
% \vskip -0.05in
\renewcommand\arraystretch{0.6}
\centering
  \caption{Ablation results analyzing the contributions of three key components: AlphaNet, MHCB, and loss function on Set11.}
  \label{tab:5}
  \vspace{-0.12in}

  % \begin{tabular}{c|ccc|c}
  \begin{tabular}{p{0.55cm}<{\centering} | p{0.88cm}<{\centering}  p{0.6cm}<{\centering}  p{0.52cm}<{\centering} | p{0.59cm}<{\centering}  p{0.59cm}<{\centering}  p{0.59cm}<{\centering}  p{0.59cm}<{\centering}}
    \toprule

    \rowcolor{cvprgray} \small{Case}&\footnotesize{AlphaNet}&\footnotesize{MHCB}&\footnotesize{LOSS}&\small{0.10}&\small{0.25}&\small{0.40}&\small{Avg.}\\
    \midrule
    (a) & \scriptsize{\XSolidBrush} &\scriptsize{\XSolidBrush} & \scriptsize{\XSolidBrush}& \small{31.45}& \small{36.40}& \small{39.72}&\small{35.86}\\
    % \multirow{7}*{\small{\footnotesize{NL-CSN}}} & $\checkmark $ & & & 27.51& 31.89& 35.52\\
  (b)  & \scriptsize{\XSolidBrush} &\checkmark & \checkmark & \small{31.68}& \small{36.64}& \small{39.86}&\small{36.06}\\
    % \cline{2-5}
  (c)  & \checkmark & \scriptsize{\XSolidBrush}  & \checkmark & \small{31.73}& \small{36.69}& \small{39.90}&\small{36.10}\\
  (d)  & \checkmark & $\checkmark $ & \scriptsize{\XSolidBrush} &\small{31.78}& \small{36.76}& \small{40.00}&\small{36.17}\\
  (e)  & \checkmark & $\checkmark $ & $\checkmark $ &\small{\textbf{31.86}}& \small{\textbf{36.87}}& \small{\textbf{40.08}}&\small{\textbf{36.27}}\\
  \bottomrule
\end{tabular}
\label{tab:5}
\vskip -0.22in
\end{table}

\begin{table}[b]
\vskip -0.16in
\renewcommand\arraystretch{0.6}
\centering
  \caption{Ablation results examining two subcomponents within MHCB on Set11. (CA: channel attention, SA: spatial attention)}
  \label{tab:6}
  \vspace{-0.12in}

  % \begin{tabular}{c|ccc|c}
  \begin{tabular}{p{1.03cm}<{\centering} | p{0.65cm}<{\centering}  p{0.65cm}<{\centering} |  p{0.75cm}<{\centering}  p{0.75cm}<{\centering}  p{0.75cm}<{\centering}  p{0.75cm}<{\centering}}
    \toprule

  \rowcolor{cvprgray}  \footnotesize{Module}&\footnotesize{CA}&\footnotesize{SA}&\small{0.10}&\small{0.25}&\small{0.40}&\small{Avg.}\\
    \midrule
    \multirow{4}*{\footnotesize{MHCB}} & \scriptsize{\XSolidBrush} &\scriptsize{\XSolidBrush} & \small{31.73}& \small{36.69}& \small{39.90}& \small{36.10}\\
    % \multirow{7}*{\small{\footnotesize{NL-CSN}}} & $\checkmark $ & & & 27.51& 31.89& 35.52\\
    % \cline{2-5}
    & \scriptsize{\XSolidBrush} & $\checkmark $ & \small{31.81}& \small{36.84}& \small{40.03}& \small{36.22}\\
    & \checkmark & \scriptsize{\XSolidBrush}&\small{31.77}& \small{36.72}& \small{39.95}& \small{36.14}\\
    & \checkmark & $\checkmark $ & \small{\textbf{31.86}}& \small{\textbf{36.87}}& \small{\textbf{40.08}}& \small{\textbf{36.27}}\\
  \bottomrule
\end{tabular}
\label{tab:6}
\end{table}

\begin{table}[b]
\vskip -0.12in
\renewcommand\arraystretch{0.6}
\centering
  \caption{Ablation results examining two subcomponents within the proposed loss function on Set11. (Data: $\mathcal{L}_{data}$, Div: $\mathcal{L}_{div}$)}
  \label{tab:7}
  \vspace{-0.12in}

  % \begin{tabular}{c|ccc|c}
  \begin{tabular}{p{1.03cm}<{\centering} | p{0.65cm}<{\centering}  p{0.65cm}<{\centering} |  p{0.75cm}<{\centering}  p{0.75cm}<{\centering}  p{0.75cm}<{\centering}  p{0.75cm}<{\centering}}
    \toprule

  \rowcolor{cvprgray}  \footnotesize{Module}&\footnotesize{Data}&\footnotesize{Div}&\small{0.10}&\small{0.25}&\small{0.40}&\small{Avg.}\\
    \midrule
    \multirow{4}*{\footnotesize{LOSS}} & \scriptsize{\XSolidBrush} &\scriptsize{\XSolidBrush} & \small{31.78}& \small{36.76}& \small{40.00}& \small{36.17}\\
    % \multirow{7}*{\small{\footnotesize{NL-CSN}}} & $\checkmark $ & & & 27.51& 31.89& 35.52\\
    % \cline{2-5}
    & \scriptsize{\XSolidBrush} & $\checkmark $ & \small{31.81}& \small{36.80}& \small{40.05}& \small{36.22}\\
    & \checkmark & \scriptsize{\XSolidBrush}& \small{31.83}& \small{36.84}& \small{40.06}& \small{36.24}\\
    & \checkmark & $\checkmark $ & \small{\textbf{31.86}}& \small{\textbf{36.87}}& \small{\textbf{40.08}}& \small{\textbf{36.27}}\\
  \bottomrule
\end{tabular}
\label{tab:7}
\end{table}

% We also compare the reconstruction speed between different CS methods. Specifically, we test all CS networks on the same platform with 3.4 GHz Intel i7 CPU plus NVIDIA GTX 3090 GPU. Table~\ref{tab:6} shows the average running time comparisons, from which we observe that the proposed IE-DUN nearly remains the same order of magnitude as the other recent CS networks. Besides, although the proposed on-site fine-tuning strategy in IE-DUN-(OSFT) introduces additional computational costs, it is usually tolerable in some CS systems. This is because compressive sensing typically serves asymmetric signal processing systems~\cite{7452635,gao2015block}, in which the encoder (\ie, sampling) is usually made computationally simple, while shifting the heavy computation burdens to the decoder (\ie, reconstruction) side. % Therefore, in many CS systems, there is usually a high tolerance for reconstruction speed.

Beyond natural images, we further extend MHC-DUN to compressive sensing MRI (CS-MRI) for reconstructing images from undersampled Fourier measurements. In this context, the sampling operator is typically defined as $\boldsymbol{\Phi} = \mathbf{SF}$, where $\mathbf{S}$ is a sub-sampling mask and $\mathbf{F}$ denotes the discrete Fourier transform. The experimental results on the brain datasets~\cite{zhangISTANetInterpretableOptimizationInspired2018}, presented in Table~\ref{tab:3}, demonstrate that the proposed MHC-DUN consistently attains superior reconstruction quality across the tested sampling rates.

Table~\ref{tab:4} reports the number of parameters, GFLOPs, and average inference time. Although MHC-DUN requires 10.81M parameters, it still runs in just 0.06s on GPU, and is faster than NesTD-Net and CPP-Net. By sharing weights across stages, MHC-DUN$^{\textbf{*}}$ shrinks to 3.68M parameters while retaining the same FLOPs, yielding the lowest latency. Overall, the proposed MHC-DUN achieves state-of-the-art reconstruction quality with favorable complexity.

%\begin{figure}[t]
%\begin{center}
%
%% \hspace{1.06in}
%% \vskip -0.13in
%\includegraphics[width=0.75in]{visual3/dudu.png}  \hspace{-0.001in}
%\includegraphics[width=0.75in]{visual3/haha.png}  \hspace{-0.001in}
%\includegraphics[width=0.75in]{visual3/haha.png}  \hspace{-0.001in}
%\includegraphics[width=0.75in]{visual3/haha.png}  \hspace{-0.001in}
%
%\end{center}
%\vskip -0.16in
%\caption{Medical images results. of on-site fine-tuning process under different learning rates (lr) for the testing image in Figure~\ref{fig:5}. (Patch size is 256$\times$256, Sampling rate is 0.10).}
%\vskip -0.23in
%\label{fig:8}
%\end{figure}

%-------------------------------------------------------------------------
\subsection{Ablation Studies and Discussions}

%%%%%%%%%%%%%%%%%% To evaluate the contribution of our proposed message transmission mechanism (including intra-phase and inter-phase transmissions), several model variants are designed, in which certain functional parts are selectively discarded or retained. Specifically,

% \textbf{Network Components: }To validate the effectiveness of the key components in MHC-DUN, we conduct ablation studies on the Set11 dataset at different sampling rates. As shown in Table 4, removing AlphaNet (case (b)) degrades performance by 0.15 dB on average, highlighting its role in enabling dynamic, collaborative gradient updates across hypotheses. Analogously, the modules MHCB and Loss function respectively yield gains of 0.14 dB and 0.16 dB on average. The above results demonstrate that each component synergistically improves reconstruction quality, with the full MHC-DUN attaining superior performance.

\textbf{Network Components: }To evaluate the key components of MHC-DUN, we perform ablation studies on Set11 across various sampling rates. As shown in Table~\ref{tab:5}, removing AlphaNet (case (b)) degrades 0.21dB on average, underscoring its role in enabling dynamic collaborative gradient updates. Analogously, MHCB (case (c)) and the loss function (case (d)) contribute average gains of 0.17dB and 0.10dB, respectively. The above results demonstrate that each component synergistically improves reconstruction quality, with the full MHC-DUN attaining superior performance.

% % . Specifically, for intra-phase message transmission, the generated gradient map (Grad) and step size map (Step) are analyzed, and for inter-phase message transmission, the blocks CNB and TRB are discussed.

\begin{table}[t]
% \vskip -0.05in
\renewcommand\arraystretch{0.6}
\centering
  \caption{Ablation analysis of the number of hypotheses (T=1, 4, 8, 16, 32) across different CS ratios on dataset Set11.}
  \label{tab:8}
  \vspace{-0.12in}

  % \begin{tabular}{c|ccc|c}
  \begin{tabular}{p{1.95cm}<{\centering} | p{0.75cm}<{\centering}  p{0.75cm}<{\centering}  p{0.75cm}<{\centering}  p{0.75cm}<{\centering}  p{0.75cm}<{\centering}}
    \toprule

    \rowcolor{cvprgray} \small{T}&\footnotesize{1}&\footnotesize{4}&\footnotesize{8}&\small{16}&\small{32}\\
    \midrule
    \small{Rate\ =\ 0.10} & \small{31.62} &\small{31.71} & \small{31.78}& \small{31.86}& \small{31.87}\\
    \small{Rate\ =\ 0.25} & \small{36.61} &\small{36.70} & \small{36.79}& \small{36.87}& \small{36.89}\\
    \small{Rate\ =\ 0.30} & \small{37.76} &\small{37.88} & \small{38.00}& \small{38.10}& \small{38.10}\\
    \small{Rate\ =\ 0.40} & \small{39.69} &\small{39.84} & \small{39.95}& \small{40.08}& \small{40.09}\\
    \midrule
    \small{\textbf{Avg.}} & \small{\textbf{36.42}} &\small{\textbf{36.53}} & \small{\textbf{36.63}}& \small{\textbf{36.72}}& \small{\textbf{36.74}}\\
  \bottomrule
\end{tabular}
\vspace{-0.22in}
\label{tab:8}
\end{table}

\textbf{MHCB Block: }We further ablate spatial and channel attentions within MHCB. As shown in Table~\ref{tab:6}, removing spatial attention results in a 0.13dB PSNR drop, highlighting its role in capturing spatial fine-grained priors, while eliminating channel attention causes a 0.05dB decrease. Besides, the absence of both leads to a cumulative 0.17dB loss, confirming that joint spatial-channel attention within MHCB is effective for exploiting intra and inter-hypothesis priors. Fig.~\ref{fig:7} presents visualizations of some spatial attention maps, illustrating that different hypotheses are adaptively integrated with varying weights.

% We ablated spatial and channel attention within MHCB to assess their contributions. As shown in Table *, removing spatial attention causes a 0.08 dB PSNR drop, while eliminating channel attention results in a 0.23 dB decrease. Removing both leads to a cumulative 0.13 dB reduction, confirming their complementary roles. These results indicate that joint spatial-channel attention is crucial for effectively leveraging intra- and inter-hypothesis priors.

\textbf{Loss Function: }The contribution of the data-fidelity term ($\mathcal{L}_{data}$) and the diversity regularizer ($\mathcal{L}_{div}$) in the proposed loss function is also evaluated. As shown in Table~\ref{tab:7}, excluding $\mathcal{L}_{data}$ causes a substantial 0.05dB PSNR drop, underscoring the necessity of measurement consistency for accurate reconstruction. Moreover, removing $\mathcal{L}_{div}$ yields an average 0.03dB decrease, validating its role in promoting complementary hypotheses. As above, the ablation results confirm that $\mathcal{L}_{data}$ and $\mathcal{L}_{div}$ are both effective and synergistic within our proposed composite loss.

\textbf{Number of Hypotheses: }The number of hypotheses (i.e., T) in our framework is analyzed. As shown in Table~\ref{tab:8}, PSNR improves from T=1 to 16, with gains of 0.11 dB, 0.10 dB, and 0.09 dB, respectively, while the advancement from 16 to 32 is negligible (i.e., 0.02 dB). As above, we set T=16 in our multi-hypothesis model. Fig.~\ref{fig:6} illustrates the visualizations of multiple hypotheses alongside their corresponding step size maps, demonstrating both the diversity among hypotheses and the adaptivity of the gradient updates.

\vskip -0.18in
\section{Conclusion}

This work designs a novel MHC-DUN for image CS reconstruction. By explicitly modeling and jointly optimizing multiple candidate solutions, MHC-DUN effectively addresses the inherent uncertainty of ill-posed inverse problems. The framework integrates adaptive step-size prediction via AlphaNet, multi-hypothesis collaborative proximal mapping, and a composite loss function that balances measurement fidelity, hypothesis diversity, and reconstruction quality. Extensive experiments confirm that MHC-DUN outperforms state-of-the-art CS methods.

\section{Acknowledgement}

This work was supported in part by the National Key R\&D Program of China (2025ZD1601300), the National Natural Science Foundation of China (NSFC) under grant 62302128, 62272128 and 62402138, the Fundamental and Interdisciplinary Disciplines Breakthrough Plan of the Ministry of Education of China (JYB2025XDXM901), and the Suzhou Key Core Technology Project under grant number SYG2025118.

{
    \small
    \bibliographystyle{ieeenat_fullname}
    \bibliography{main}
}


\end{document}